\documentclass[aps,prx,showkeys,reprint]{revtex4-2}

\usepackage[utf8]{inputenc}
\usepackage[T1]{fontenc}
\usepackage{amsmath}
\usepackage{amssymb}
\usepackage{amsfonts}
\usepackage{upgreek}
\usepackage{mathtools}
\usepackage{chemformula}
\usepackage{float}
\usepackage{graphicx}
\usepackage{tabularx}
\usepackage[labelfont=bf]{caption}

\usepackage[hidelinks]{hyperref}

\begin{document}

\title{Lifelong Machine Learning Potentials}

\author{Marco Eckhoff}
\email{marco.eckhoff@phys.chem.ethz.ch}
\author{Markus Reiher}
\email{mreiher@ethz.ch}
\affiliation{ETH Z\"urich, Departement Chemie und Angewandte Biowissenschaften, Vladimir-Prelog-Weg 2, 8093 Z\"urich, Switzerland.}

\date{May 11, 2023}

\begin{abstract}

Machine learning potentials (MLPs) trained on accurate quantum chemical data can retain the high accuracy, while inflicting little computational demands. On the downside, they need to be trained for each individual system. In recent years, a vast number of MLPs has been trained from scratch because learning additional data typically requires to train again on all data to not forget previously acquired knowledge. Additionally, most common structural descriptors of MLPs cannot represent efficiently a large number of different chemical elements. In this work, we tackle these problems by introducing element-embracing atom-centered symmetry functions (eeACSFs) which combine structural properties and element information from the periodic table. These eeACSFs are a key for our development of a lifelong machine learning potential (lMLP). Uncertainty quantification can be exploited to transgress a fixed, pre-trained MLP to arrive at a continuously adapting lMLP, because a predefined level of accuracy can be ensured. To extend the applicability of an lMLP to new systems, we apply continual learning strategies to enable autonomous and on-the-fly training on a continuous stream of new data. For the training of deep neural networks, we propose the continual resilient (CoRe) optimizer and incremental learning strategies relying on rehearsal of data, regularization of parameters, and the architecture of the model.

\end{abstract}


\maketitle

%%%%%%%%%%%%%%%%%%%%%%%%%%%%%%%%%%%%%%%%%%%%%%%%%%%%%%%%%%%%%%%%%%%%%%%%%%%%%%%%%%%%%%%%%%%%%%%%%%%%%%%%
\section{Introduction}
%%%%%%%%%%%%%%%%%%%%%%%%%%%%%%%%%%%%%%%%%%%%%%%%%%%%%%%%%%%%%%%%%%%%%%%%%%%%%%%%%%%%%%%%%%%%%%%%%%%%%%%%

For the prediction and understanding of the properties and reactivity of atomistic systems, the knowledge of the potential energy surface is inevitable. The potential energy surface is given by the electronic energy as a function of the nuclear positions and can be obtained from electronic structure methods such as density functional theory (DFT) or approximate wave-function theories \cite{Cramer2013, Jensen2017}. These methods are generally applicable and achieve accurate results for many systems, but even state-of-the-art approaches still lead to high computational demands in extended simulations \cite{Burke2012, Riplinger2016, Das2019}. Instead of explicitly calculating the electronic energy for each nuclear conformation in Born-Oppenheimer approximation, empirical force fields rely on approximate (simple) analytical expressions of the potential energy surface. Hence, they avoid the explicit integration of quantum mechanical equations and thereby enabling efficient atomistic simulations. However, force field expressions are of limited accuracy and they are typically not universal for all chemical systems \cite{vanDuin2001, Piana2011, Friederich2021, Behler2021a}.

By contrast, machine learning potentials (MLPs) \cite{Behler2016, Bartok2017, Deringer2019, Noe2020, Westermayr2021a, Kaeser2023} can preserve the high accuracy of electronic structure methods, but at low computational cost comparable to that of force fields. MLPs are not based on physical approximations, but rely on very flexible mathematical expressions to obtain an analytical potential energy surface. The parameters of these expressions are trained on electronic structure reference data including chemical structures and their respective energies and atomic forces. While first-generation MLPs were only applicable to low-dimensional systems \cite{Blank1995}, the introduction of second-generation MLPs in form of high-dimensional neural network potentials (HDNNPs) \cite{Behler2007, Behler2017, Behler2021} has facilitated atomistic simulations of systems with tens of thousands of atoms on nanosecond timescales with an accuracy similar to that of first-principles methods. In recent years, various other types of MLPs have been proposed such as neural network potentials \cite{Schuett2017, Unke2019, Unke2021, Batatia2022}, Gaussian approximation potentials \cite{Bartok2010}, moment tensor potentials \cite{Shapeev2016}, and many more. Second-generation MLPs rely on the locality of the major part of atomic interactions \cite{Behler2021}. However, third-generation MLPs add the description of long-range interactions \cite{Bartok2010, Artrith2011} and fourth-generation MLPs even non-local interactions \cite{Ghasemi2015, Ko2021} beyond the applied cutoff sphere for atomic interactions.

Reference atomistic structures need to represent sufficiently well the conformation space to be explored in subsequent simulations. The reason is that MLPs are designed for interpolation of learned atomic interactions but the extrapolation capability far beyond the trained atomic environment space is limited \cite{Behler2017, Deringer2019}. For example, an MLP trained on water will fail for gaseous dihydrogen and dioxygen. A systematic construction approach is not feasible for most systems because of the enormous conformation space that is typically accessible for an atomistic system, especially when one considers various large-amplitude motions or chemical reactions. To obtain reasonable reference conformations, iterative active learning protocols have been devised \cite{Artrith2012, Behler2015, Podryabinkin2017, Bernstein2019, Eckhoff2021}. In such protocols, preliminary MLPs are validated in atomistic simulations to identify important, but yet unknown conformations. These conformations are recalculated with the reference method and then added to the reference data to train an improved MLP. However, it is difficult to ensure that a reference data set is complete even for a specific application because some conformations may occur only very rarely. Also, different scientific targets and purposes may, for the same system, highlight different regions of its conformation space and, therefore, require different reference conformations.

Current MLPs learn in isolation, i.e., a model is trained on predefined data and is subsequently applied in simulations without exploiting for future MLPs the knowledge already learned. Despite the general functional form, MLPs can therefore only be applied reliably to a specific conformation space---which can span various chemical formulae and configurations---and the transferability significantly beyond their learned atomic environment space is limited \cite{Friederich2021, Behler2021a}. As a consequence, a vast number of single-purpose MLPs has been constructed \cite{Morawietz2016, Hellstroem2016, Cheng2019, Westermayr2019, Amabilino2019, Eckhoff2019, Eckhoff2020a, Eckhoff2021a, Zuo2020}. Recent work has attempted to overcome this limitation by leveraging very large and broad training data sets \cite{Smith2017, Zhang2022}, but future applications may still require different reference data.

To achieve a general-purpose solution, MLPs need to be adaptable to learn new chemical systems based on already acquired knowledge. Obviously, the training data sets of present-day MLPs can be simply extended to include additional systems in active learning procedures. However, the training process of new or extended data sets is often started from scratch with randomly initialized MLP parameters. Even if previously obtained parameters are used, the training typically does not differentiate between new and old data points. Hence, all data points need to be trained simultaneously from a stationary batch of data every time new data is added. Consequently, learning larger data sets becomes increasingly inefficient as the amount of repeated training of old data grows. This trend is opposite to how the human brain operates which efficiently learns additional knowledge based on already acquired expertise \cite{Hassabis2017, Chen2018, Parisi2019}.

This ability of incremental learning is referred to as lifelong learning or continual learning in the field of machine learning \cite{Chen2018, Parisi2019} and it is also desirable for MLPs. Lifelong machine learning means that the model continually acquires and fine-tunes knowledge. For MLPs it will be a single-incremental-task scenario because new chemical structures are sequentially added but the energy needs to be distinguished among all encountered structures \cite{Maltoni2019}. Lifelong learning is a challenge for machine learning because incremental learning from a continuous stream of information typically results in catastrophic forgetting and interference \cite{Grossberg2013, Goodfellow2015}. The MLP parameters are adapted to the new data, while the memory about the old data fades away.

In principle, building on prior knowledge could even reduce the necessary amount of training data and more complex tasks can be learned more efficiently. According to complementary learning systems theory \cite{Parisi2019}, learning is based on episodic memory and generalization. While the episodic memory learns arbitrary information fast, the generalization is a slow learning process to structure the knowledge. Therefore, the episodic memory is similar to the training data set, while the MLP parameters are trained to generalize the data. Consequently, the parameters need to be plastic to integrate new knowledge, but, at the same time, they need to be stable to avoid forgetting \cite{Hassabis2017, Chen2018, Parisi2019}. In recent years, several algorithms have been developed to reach this stability-plasticity balance and to mitigate catastrophic forgetting; examples are ExStream \cite{Hayes2019}, gradient episodic memory \cite{Lopez-Paz2017}, synaptic intelligence \cite{Zenke2017}, AR1 \cite{Maltoni2019}, progressive neural networks \cite{Rusu2022}, and growing dual-memory \cite{Parisi2018}. These approaches rely on rehearsal of selected training data, regularization of machine learning parameters, and/or the architecture of the machine learning model.

The situation becomes more complex for a general-purpose or universal MLP as it would be further hindered by the unfavorable scaling in computational cost with the number of chemical elements. This feature plagues most common MLP descriptors \cite{Behler2021a, Unke2021a, Kulik2022} such as atom-centered symmetry functions (ACSFs) \cite{Behler2011}, smooth overlap of atomic positions \cite{Bartok2013}, and many more \cite{Musil2021, Gugler2022}---with exception of graph representations \cite{Scarselli2009, Gilmer2017, Chen2022}. 
The reason is a rapid increase in the number of structural descriptors since these are constructed independently of one another for each element combination. Some studies have attempted to overcome this problem by combining element and structure information in the structural descriptor through the inclusion of a single additional element-dependent weight \cite{Artrith2017, Rostami2018, Gastegger2018}. Despite some successful applications, employing, for example, the atomic number as in weighted ACSFs \cite{Gastegger2018} is restricted to elements of similar atomic numbers. Otherwise, the contributions of light atoms, such as H, are obscured by contributions of heavy atoms, such as iodine (e.g., factor $53\cdot53$ in angular contributions).

In this work, we exploit additional weight terms based on different properties for separate ACSFs that can reduce bias toward specific elements and increase the discriminability of the descriptor representation for different conformations. Chemical intuition is based on regularities in molecular structure and on trends in the periodic table. To exploit these trends, we use information about the period and the group \cite{Faber2018}, or more precisely, the position in the s-, p-, d-, and f-block instead of the atomic number. For instance, all halogens lead to a similar type of bonding, while the bond length increases in higher periods.

Furthermore, we propose that uncertainty quantification can enable the application of adaptable MLP parameters. Adaptable parameters require that the reliability of an MLP is probed on the fly to avoid full validation of each MLP training extension. As long as the error of the MLP is below a tolerance in production calculations, little variations with the learning process will not significantly affect the results. Moreover, MLPs with uncertainty quantification are able to report warnings in case of unknown conformations, even if these are within the general training space. Therefore, the uncertainty quantification can be employed to estimate the transferability of the MLP, with low uncertainty meaning high transferability. To obtain uncertainty quantification, for example, for HDNNPs, some studies have applied an ensemble approach \cite{Peterson2017, Smith2018, Smith2019, Devereux2020, Imbalzano2021}. A small ensemble of HDNNPs is then trained differently and independently on the same reference data. The deviations between the predictions of the individual HDNNPs can be employed as a proxy to the prediction error due to the high flexibility of the neural networks. Moreover, taking the average of the ensemble improves the accuracy of the representation. We note that such uncertainty measures should accompany any sort of modeling approach in atomistic simulations \cite{Reiher2022}, although their implementation has only started recently.

Consequently, this work introduces element-embracing atom-centered symmetry functions (eeACSFs) to overcome the unfavorable scaling with the number of elements up to an arbitrary number. In combination with uncertainty quantification and continual learning strategies, the concept of a lifelong machine learning potential (lMLP) is proposed, which can be trained in a rolling fashion by a continuous stream of new data. To perform this challenging training process, we present the new continual resilient (CoRe) optimizer which is an adaptive method for stochastic first-order iterative optimization. CoRe is applied in combination with our lifelong training strategies which provide an adaptive selection of employed training data, including reduction of the data set size and removal of doubtful data. We demonstrate the performance of our lMLP concept for a data set including 42 different S$_\text{N}$2 reactions and thereby ten different chemical elements. Although our work on the lMLP concept rests on a second-generation HDNNP representation, we emphasize that the concept can also be applied for a different base model. Furthermore, the CoRe optimizer and the lifelong adaptive data selection can be employed in training of machine learning models beyond lMLPs and contribute to the development of lifelong machine learning to mitigate catastrophic forgetting.

This work is organized as follows: In Section \ref{sec:Methods} we summarize the HDNNP method and introduce eeACSFs, the CoRe optimizer, lifelong adaptive data selection, uncertainty quantification, and lMLPs. After presenting the computational details in Section \ref{sec:Comp_Details}, Section \ref{sec:Results} starts with a description of the reference data. Section \ref{sec:Results} continues with a performance assessment of the eeACSF representation, a comparison between the results of CoRe and those of other optimizers, performance tests of the lifelong training strategies, and a validation of the uncertainty quantification in potential energy surface predictions. This work ends with a conclusion in Section \ref{sec:Conclusion}.

%%%%%%%%%%%%%%%%%%%%%%%%%%%%%%%%%%%%%%%%%%%%%%%%%%%%%%%%%%%%%%%%%%%%%%%%%%%%%%%%%%%%%%%%%%%%%%%%%%%%%%%%
\section{Methods}\label{sec:Methods}
%%%%%%%%%%%%%%%%%%%%%%%%%%%%%%%%%%%%%%%%%%%%%%%%%%%%%%%%%%%%%%%%%%%%%%%%%%%%%%%%%%%%%%%%%%%%%%%%%%%%%%%%

%%%%%%%%%%%%%%%%%%%%%%%%%%%%%%%%%%%%%%%%%%%%%%%%%%%%%%%%%%%%%%%%%%%%%%%%%%%%%%%%%%%%%%%%%%%%%%%%%%%%%%%%
\subsection{High-Dimensional Neural Network Potential with Standardization}
%%%%%%%%%%%%%%%%%%%%%%%%%%%%%%%%%%%%%%%%%%%%%%%%%%%%%%%%%%%%%%%%%%%%%%%%%%%%%%%%%%%%%%%%%%%%%%%%%%%%%%%%

For a system containing $N_\mathrm{elem}$ elements and $N_\mathrm{atom}^m$ atoms of element $m$, the second-generation HDNNP energy \cite{Behler2007, Behler2017, Behler2021} is given by a sum of atomic energy contributions $E_{\mathrm{atom},n}^m$ of every atom $n$:
\begin{align}
E=\sum_{m=1}^{N_\mathrm{elem}}\sum_{n=1}^{N_\mathrm{atom}^{m}}E_{\mathrm{atom},n}^m\ .
\end{align}
Each atomic energy contribution is obtained from a feed-forward neural network,
\begin{align}
\begin{split}
&E_{\mathrm{atom},n}^m=b_1^{m,3}+\sum_{\lambda=1}^{n_2}a_{\lambda1}^{m,23}\cdot f^2\Bigg\{b_\lambda^{m,2}+\sum_{\kappa=1}^{n_1}a_{\kappa\lambda}^{m,12}\\
&\cdot f^1\Bigg[b_\kappa^{m,1}+\sum_{i=1}^{n_G}a_{i\kappa}^{m,01}\cdot\alpha_i^m\Bigg(G_{n,i}^m-\beta_i^m\Bigg)\Bigg]\Bigg\}\ .
\end{split}
\end{align}
The weights $a$ and $b$ are trained individually for each element $m$ according to energies and atomic forces of a training data set containing multiple chemical structures. We note that the number of hidden layers can differ from the case shown, which is for two hidden layers with $n_1$ and $n_2$ neurons each. A vector of local structural descriptors $\mathbf{G}_n^m$ of dimension $n_G$, which is explained in the next section, is given as input to this atomic neural network. We propose the activation function $f(x)=1.59223\cdot\tanh(x)$, which is discussed in Section S1.1 of the Supporting Information along with a tailored weight initialization scheme \cite{Eckhoff2020a, Eckhoff2020b}. The convergence advantage and accuracy increase of this activation function and weight initialization compared to a hyperbolic tangent and a non-tailored weight initialization is shown in Table S3 and Figures S1 (a), S1 (b), and S2 in the Supporting Information.

In contrast to the original HDNNP method \cite{Behler2007} the trainable weights $\alpha_i^m$ and $\beta_i^m$ are introduced for standardization of the structural descriptor input $G_{n,i}^m$. For this purpose, the structural descriptor is shifted by $\beta_i^m$, which is initialized by the mean of the descriptor values $G_{n,i}^m$ in the initial training data set for all atoms $n$ of the respective element $m$. The result is scaled by $\alpha_i^m$, which is initialized by the inverse standard deviation of the respective descriptor values in the initial training data. In this way, the weights $a_{i\kappa}^{m,01}$ of the input layer are multiplied with values, which are centered around zero and show a standard deviation of one for the initial training data. This standardization can improve the training performance and is adjustable in case of additional training data. We note that the weight initialization can be restricted to values inside a certain interval to avoid numerical issues. The weight pair $a_{i\kappa}^{m,01}$ and $\alpha_i^m$ may be combined, but we treat them separately for technical reasons during the optimization. 

%%%%%%%%%%%%%%%%%%%%%%%%%%%%%%%%%%%%%%%%%%%%%%%%%%%%%%%%%%%%%%%%%%%%%%%%%%%%%%%%%%%%%%%%%%%%%%%%%%%%%%%%
\subsection{Element-Embracing Atom-Centered Symmetry Functions}
%%%%%%%%%%%%%%%%%%%%%%%%%%%%%%%%%%%%%%%%%%%%%%%%%%%%%%%%%%%%%%%%%%%%%%%%%%%%%%%%%%%%%%%%%%%%%%%%%%%%%%%%

HDNNPs usually employ vectors of many-body atom-centered symmetry functions \cite{Behler2011} to represent the local atomic environments. These descriptors fulfill the translational, rotational, and permutational invariances of the potential energy surface. They depend on distances and angles of all neighboring atoms which are inside a cutoff sphere of radius $R_\mathrm{c}$. This radius needs to be sufficiently large to account for all relevant interactions. Since no connectivities are used, the HDNNP is able to describe chemical reactions. The dimensionality of the input vector does not depend on the individual atomic environment---a requirement to obtain generally applicable atomic neural networks. Conventional ACSF vectors are constructed in such a way that each ACSF represents only all interactions between a specific chemical element pair or triple. This construction, however, leads to an unfavorable scaling in the number of descriptors with respect to the number of elements. 

For this reason, we introduce element-embracing atom-centered symmetry functions (eeACSFs) which explicitly depend on the element information $H$ from the periodic table (see next paragraph). Similar to ACSFs, there are two types of eeACSFs. The radial eeACSFs,
\begin{align}
\begin{split}
&G_{n,i}^{\mathrm{rad}}=\Bigg[\frac{1}{H_{\mathrm{max},i}^\mathrm{rad}}\sum_{j\neq n}^{N_\mathrm{atom}}H_{i,j}^\mathrm{rad}\exp\left(-\eta_i^\mathrm{rad}\cdot R_{nj}^2\right)\\
&\cdot f_\mathrm{c}\left(R_{nj}\right)\Bigg]^{\tfrac{1}{2}}\ ,
\end{split}
\end{align}
are a function of the distances $R_{nj}$ between the central atom $n$ and the neighboring atoms $j$. The angular eeACSFs,
\begin{align}
\begin{split}
&G_{n,i}^{\mathrm{ang}}=\Bigg\{\frac{2^{-\zeta_i}}{H_{\mathrm{max},i}^\mathrm{ang}}\sum_{j\neq n}^{N_\mathrm{atom}}\sum_{k\neq n,j}^{N_\mathrm{atom}}H_{i,jk}^\mathrm{ang}\left[1+\lambda_i\cos\left(\theta_{njk}\right)\right]^{\zeta_i}\\
&\cdot\exp\left[-\eta_i^\mathrm{ang}\left(R_{nj}^2+R_{nk}^2\right)\right]\cdot f_\mathrm{c}\left(R_{nj}\right)\cdot f_\mathrm{c}\left(R_{nk}\right)\Bigg\}^{\tfrac{1}{2}}\ ,
\end{split}
\end{align}
depend in addition on the angle $\theta_{njk}$ between atom $n$ and the two neighbors $j$ and $k$. Different values for the parameters $\eta_i^\mathrm{rad}\geq0$, $\eta_i^\mathrm{ang}\geq0$, $\lambda_i=\pm1$, and $\xi_i\geq1$ of each eeACSF $i$ eventually produce a structural fingerprint of the atomic environment of atom $n$. The cutoff function,
\begin{align}
f_\mathrm{c}\left(R_{nj}\right)=\begin{cases}\exp\left[1-\left(1-\frac{R_{nj}^2}{R_\mathrm{c}^2}\right)^{-1}\right]&\mathrm{for}\ R_{nj}<R_\mathrm{c}\\
0&\mathrm{otherwise}\end{cases}\ ,
\end{align}
damps the contributions smoothly to zero beyond the cutoff radius $R_\mathrm{c}$. The advantage of this cutoff function is that also the derivatives of all orders are zero at $R_\mathrm{c}$, which is beneficial for the calculation of forces, normal modes, and so forth. In contrast to ACSFs, a square root is applied to eeACSFs to mitigate a too strong effect of the number of neighbors on the eeACSF value. Such a strong effect can be observed for training data including very different molecule sizes and particle densities, and it can decrease parametrization performance due to a too broad range of input values.

As element descriptors we propose $n$ for the period number of the element in the periodic table, $m$ for the group number in the s- and p-block (main group 1 to 8), and $d$ for the group number in the d-block. Main group elements obey $d=0$ and d-block elements $m=0$. A special case is helium with $m=8$. The values of these descriptors for the neighboring atom $j$ are used in the element-dependent term $H_{i,j}^\mathrm{rad}$ of the radial eeACSFs,
\begin{align}
\begin{split}
&H_{i,j}^\mathrm{rad}\in\left\{1,\ n_j,\ m_j,\ d_j,\ \overline{n}_j\coloneqq X-n_j,\right.\\
&\left.\overline{m}_j\coloneqq9-m_j,\ \overline{d}_j\coloneqq11-d_j\right\}\ . 
\end{split}
\end{align}
For the construction of radial eeACSFs without element dependence, $H_{i,j}^\mathrm{rad}=1$ can be applied. In this work, the maximum period is set to $X-1=5$, i.e., elements up to xenon are considered. The descriptors $\overline{n}$, $\overline{m}$, and $\overline{d}$ are employed to balance contributions of light and heavy elements as well as of elements with few and many valence electrons. For main group elements $\overline{d}=0$ is used and for d-block elements $\overline{m}=0$. For elements of higher periods, the group in the f-block can be implemented in the same way as for the d-block and $X-1$ can be set to 7. To keep the contribution of each interaction to a value between 0 and 1, the radial eeACSF is divided by $H_{\mathrm{max},i}^\mathrm{rad}$ which is the maximum possible value of $H_{i,j}^\mathrm{rad}$.

The element-dependent terms $H_{i,jk}^\mathrm{ang}$ of the angular eeACSFs are calculated as linear combinations,
\begin{align}
H_{i,jk}^\mathrm{ang}=\left|H_{i,j}^\mathrm{rad}+\gamma_iH_{i,k}^\mathrm{rad}\right|+C_i\ ,
\end{align}
with $\gamma_i=\pm1$ and
\begin{align}
C_i=\begin{cases}0&\mathrm{for}\ \gamma_i=1\lor H_{i,j}^\mathrm{rad}=H_{i,k}^\mathrm{rad}=0\\
1&\mathrm{otherwise}\end{cases}\ .
\end{align}
As a consequence, the element-dependent prefactor of the angular eeACSF is defined as 
\begin{align}
H_{\mathrm{max},i}^\mathrm{ang}=\begin{cases}2H_{\mathrm{max},i}^\mathrm{rad}&\mathrm{for}\ \gamma_i=1\\
H_{\mathrm{max},i}^\mathrm{rad}&\mathrm{otherwise}\end{cases}\ .
\end{align}

In conclusion, for systems with/without d-block elements five/seven different $H_{i,j}^\mathrm{rad}$ and nine/eleven $H_{i,jk}^\mathrm{ang}$ terms are required. In combination with typically around five $\eta_i^\mathrm{rad}$, two $\eta_i^\mathrm{ang}$, two $\lambda_i$, and three $\zeta_i$ parameter values, the descriptor vector will consist of 25/35 radial and 108/132 angular eeACSFs independent of the number of elements. By contrast, the number of radial ACSFs is proportional to the number of elements and the number of angular ACSFs scales with $\sum_{m=1}^{N_\mathrm{elem}}m$. For example, in the case of four elements, 20 radial and 120 angular ACSFs are obtained for the same number of parameter values. Hence, at around four elements is the break-even point of computational cost, since the additional effort for determining $H_{i,j}^\mathrm{rad}$ and $H_{i,jk}^\mathrm{ang}$ from the listed element-dependent values is small. We note that the number of different parameters values can affect the resolution of the representation. The computational bottleneck in the descriptor calculation during MLP applications is the determination of the derivatives as a function of the atomic positions to obtain the atomic forces. Since no additional position-dependent properties are included in eeACSFs compared to ACSFs, the computational cost per eeACSF is similar to that of ACSFs.

%%%%%%%%%%%%%%%%%%%%%%%%%%%%%%%%%%%%%%%%%%%%%%%%%%%%%%%%%%%%%%%%%%%%%%%%%%%%%%%%%%%%%%%%%%%%%%%%%%%%%%%%
\subsection{Training of High-Dimensional Neural Network Potentials}
%%%%%%%%%%%%%%%%%%%%%%%%%%%%%%%%%%%%%%%%%%%%%%%%%%%%%%%%%%%%%%%%%%%%%%%%%%%%%%%%%%%%%%%%%%%%%%%%%%%%%%%%

To optimize the weights of the atomic neural networks with respect to potential energies $E^{\mathrm{ref},r}$ and Cartesian atomic force components $F_{\alpha,n}^{\mathrm{ref},r}$ of reference conformations $r$, a loss function is defined:
\begin{align}
\begin{split}
&L^t=\frac{q^2}{N_\mathrm{conf}}\sum_{r=1}^{N_\mathrm{conf}}\left(\frac{E^{r,t}-E^{\mathrm{ref},r}}{N_\mathrm{atom}^r}\right)^2+\left(3\sum_{r=1}^{N_\mathrm{conf}}N_\mathrm{atom}^r\right)^{-1}\\
&\cdot\sum_{r=1}^{N_\mathrm{conf}}\sum_{n=1}^{N_\mathrm{atom}^r}\sum_{\alpha=x,y,z}\left(F_{\alpha,n}^{r,t}-F_{\alpha,n}^{\mathrm{ref},r}\right)^2\ .
\end{split}\label{eq:loss}
\end{align}
The atomic force component is the negative gradient of the energy with respect to the Cartesian coordinate $\alpha_n^r={x_n^r,y_n^r,z_n^r}$ of atom $n$ of conformation $r$,
\begin{align}
F_{\alpha,n}^r=-\frac{\partial E^r}{\partial \alpha_n^r}\ .
\end{align}
A set of $N_\mathrm{conf}$ conformations is trained simultaneously in each training epoch $t$ to accelerate the optimization and reduce overfitting of single data points. Since the HDNNP prediction of a conformation can be dependent on the atomic neural networks of different chemical elements, these networks are also trained simultaneously. To balance the contributions of energies and forces to the loss function, the hyperparameter $q$ is used. This hyperparameter needs to be chosen with care as it can significantly affect the training performance. To optimize the weights, in each training epoch the gradient of the loss function with respect to the weights,
\begin{align}
w_\xi\coloneqq w_{\chi,\kappa\lambda}^{m,\mu\nu}\in\left\{a_{\kappa\lambda}^{m,\mu\nu},\ b_{\lambda}^{m,\nu},\ \alpha_{\kappa}^{m},\ \beta_{\kappa}^{m}\right\}\ ,
\end{align}
is calculated, where $\xi$ is a unique index of each weight.

%%%%%%%%%%%%%%%%%%%%%%%%%%%%%%%%%%%%%%%%%%%%%%%%%%%%%%%%%%%%%%%%%%%%%%%%%%%%%%%%%%%%%%%%%%%%%%%%%%%%%%%%
\subsection{Continual Resilient (CoRe) Optimizer}
%%%%%%%%%%%%%%%%%%%%%%%%%%%%%%%%%%%%%%%%%%%%%%%%%%%%%%%%%%%%%%%%%%%%%%%%%%%%%%%%%%%%%%%%%%%%%%%%%%%%%%%%

To improve the training process, we developed the continual resilient (CoRe) optimizer which aims to combine the robustness of resilient backpropagation (RPROP) \cite{Riedmiller1993, Riedmiller1994} with the performance of the Adam optimizer \cite{Kingma2015}. Moreover, we introduce adaptive decay rates of moving averages of the loss function gradients, plasticity factors of the weights obtained from an importance score, and weight decays bounding the weight values to increase the convergence speed and final accuracy beyond state-of-the-art optimizers.

CoRe is a first-order gradient-based optimizer. It employs individual adaptive learning rates for each weight $w_\xi$, which depend on optimization history. The algorithm is intended for stochastic iterative optimizations, i.e., subsamples of the batch of training data are used in each training epoch. Thus, computational efficiency benefits of stochastic gradient decent (SGD) \cite{Robbins1951} can be exploited.

In the spirit of the Adam optimizer, exponential moving averages of the gradient,
\begin{align}
g_\xi^\tau=\beta_1^\tau\cdot g_\xi^{\tau-1}+\left(1-\beta_1^\tau\right)\frac{\partial L^t}{\partial w_\xi^t}\ ,\label{eq:g}
\end{align}
and the squared gradient,
\begin{align}
h_\xi^\tau=\beta_2\cdot h_\xi^{\tau-1}+\left(1-\beta_2\right)\left(\frac{\partial L^t}{\partial w_\xi^t}\right)^2\ ,\label{eq:h}
\end{align}
with decay rates $\beta_1^\tau,\beta_2\in[0,1)$ are activated in the computation of the individual adaptive learning rates of each weight $w_\xi^t$. $\tau$ is the optimization step counter for each weight, which can be different from the training epoch $t$. For example, for HDNNPs a training data subsample of a training epoch may not contain every chemical element of the entire training data set leading to weight updates only for some of the atomic neural networks in the respective training epoch. In this way, the HDNNP model already provides an architectural strategy to mitigate catastrophic forgetting. We note that Equations (\ref{eq:g}) and (\ref{eq:h}) represent the case of minimization, while for maximization the sign of the loss function derivative with respect to the weight must be inverted.

In contrast to the Adam optimizer, $\beta_1$ is a function of $\tau$,
\begin{align}
\beta_1^\tau=\beta_1^\mathrm{b}+\left(\beta_1^\mathrm{a}-\beta_1^\mathrm{b}\right)\exp\left[-\left(\frac{\tau-1}{\beta_1^\mathrm{c}}\right)^2\right]\ .\label{eq:beta_1}
\end{align}
The hyperparameters $\beta_1^\mathrm{a},\beta_1^\mathrm{b}\in[0,1)$ define the initial and final values of $\beta_1$. These values are interconverted by a Gaussian with hyperparameter $\beta_1^\mathrm{c}>0$. A larger $\beta_1$ value increases the dependence on previous gradient information, that is, on the optimization history of all recent training data subsamples. Thereby, the optimization performance with respect to the entire training data set can be improved. A smaller $\beta_1$ value yields a higher dependence on the current gradient of the subsample, decreasing in some way the moment of inertia of the optimization process. The latter is beneficial for rapidly and strongly changing gradients occurring in fast convergence at the beginning of an optimization, which is started from randomly initialized weights. In conclusion, Equation (\ref{eq:beta_1}) can be employed to increase $\beta_1$ in the course of the optimization to improve the training performance.

One contribution of the weight update magnitudes of CoRe is obtained, in analogy to the Adam optimizer, according to
\begin{align}
u_\xi^\tau=\frac{g_\xi^\tau}{1-\left(\beta_1^\tau\right)^\tau}\left\{\left[\frac{h_\xi^\tau}{1-\left(\beta_2\right)^\tau}\right]^{\tfrac{1}{2}}+\epsilon\right\}^{-1}\ .
\end{align}
The moving averages $g_\xi^\tau$ and $h_\xi^\tau$ get bias corrected by $1-(\beta_1^\tau)^\tau$ and $1-(\beta_2)^\tau$, respectively, to counteract their initialization bias toward zero ($g_\xi^0,h_\xi^0=0$). The division of the two bias-corrected moving averages makes the weight update magnitudes invariant to gradient rescaling. A form of step size annealing is obtained, because a decreasing gradient value leads to a decrease of the update $u_\xi^\tau$. Therefore, for well-behaving optimizations the absolute value of $u_\xi^\tau$ typically decreases from $\pm1$ in the first optimization step $\tau=1$ towards zero. Higher values of the hyperparameter $\beta_2$ promote this annealing. The hyperparameter $\epsilon\gtrapprox0$ is added for numerical stability.

As a second contribution to the weight update, the plasticity factor,
\begin{align}
P_\xi^\tau=\begin{cases}0&\hspace{-0.1cm}\begin{array}{l}\mathrm{for}\ \tau>t_\mathrm{hist}\\
\land\ S_\xi^{\tau-1}\ \mathrm{top}\text{-}n_{\mathrm{frozen},\chi}^{m,\mu\nu}\ \mathrm{in}\ \mathbf{S}_\chi^{m,\mu\nu,\tau-1}\end{array}\\
1&\hspace{-0.1cm}\begin{array}{l}\mathrm{otherwise}\end{array}\end{cases}\ ,\label{eq:plasticity}
\end{align}
is introduced to mitigate forgetting of old information. In the initial training phase, this factor equals one. When $\tau>t_\mathrm{hist}$, with hyperparameter $t_\mathrm{hist}>0$, the plasticity factor can freeze the values of some weights by setting $P_\xi^\tau$ to zero. The selection of these weights depends on a score value $S_\xi^{\tau-1}$, which ranks the importance of the weights with regard to previously predicted loss function decrease, as will be explained in Equation (\ref{eq:score}). The score value $S_\xi^{\tau-1}$ is compared to all other score values $\mathbf{S}_\chi^{m,\mu\nu,\tau-1}$ of the same group. For HDNNPs, a group is composed by the weights within the atomic neural network of the same element $m$, with the same weight type $\chi$, and the same layer assignment $\mu\nu$. The weights belonging to the $n_\mathrm{frozen}$ highest score values in their respective group are frozen for the optimization step $\tau$. The hyperparameter $n_{\mathrm{frozen},\chi}^{m,\mu\nu}\geq0$ can be set individually for each group of weights. In this way, a regularization is established for weights with highest estimated importance.

A further contribution of the step size adjustment is adapted from RPROP,
\begin{align}
s_\xi^\tau=\begin{cases}\mathrm{min}\left(\eta_+\cdot s_\xi^{\tau-1},s_\mathrm{max}\right)&\mathrm{for}\ g_\xi^{\tau-1}\cdot g_\xi^\tau\cdot P_\xi^\tau>0\\
\mathrm{max}\left(\eta_-\cdot s_\xi^{\tau-1},s_\mathrm{min}\right)&\mathrm{for}\ g_\xi^{\tau-1}\cdot g_\xi^\tau\cdot P_\xi^\tau<0\\s_\xi^{\tau-1}&\mathrm{for}\ g_\xi^{\tau-1}\cdot g_\xi^\tau\cdot P_\xi^\tau=0\end{cases}\ .
\end{align}
In general, the step size $s_\xi^\tau$ does not depend on the magnitude of the gradient but only on its sign, yielding a robust optimization. Every time $g_\xi^\tau$ changes its sign compared to $g_\xi^{\tau-1}$, the previous optimization step probably jumped over a local minimum. Then, the previous step size $s_\xi^{\tau-1}$ was too large and needs to be decreased. If the signs of $g_\xi^\tau$ and $g_\xi^{\tau-1}$ are the same, $s_\xi^{\tau-1}$ can be increased to speed up convergence. The decrease factor $\eta_-$ and increase factor $\eta_+$, with $0<\eta_-\leq1\leq\eta_+$, are applied to obtain the current step size $s_\xi^\tau$, whereby $s_\xi^\tau$ is bounded by minimal and maximal step sizes $s_\mathrm{min},s_\mathrm{max}>0$. If $P_\xi^\tau$, $g_\xi^\tau$, and/or $g_\xi^{\tau-1}$ are zero, the step size update is omitted. The step size $s_\xi^0$ needs to be initialized determining the first optimization step size $s_\xi^1$. The value can be chosen in reasonable proportion to the initial weight values and experience has shown that the precise choice of this parameter is rather noncritical due to the fast adaption. We note that the gradient is not reset to $g_\xi^\tau=0$ for $g_\xi^{\tau-1}\cdot g_\xi^\tau<0$, in contrast to some RPROP variants which include a backtracking weight step \cite{Riedmiller1993, Igel2000}. As a consequence, the history of $g_\xi^\tau$ can be retained.

To decrease the risk of overfitting, the weight update,
\begin{align}
w_\xi^t=\left(1-d_\chi^{m,\mu\nu}\cdot\left|u_\xi^\tau\right|\cdot P_\xi^\tau\cdot s_\xi^\tau\right)w_\xi^{t-1}-u_\xi^\tau\cdot P_\xi^\tau\cdot s_\xi^\tau\ ,
\end{align}
includes a weight decay with weight decay hyperparameter $d_\chi^{m,\mu\nu}\in[0,(s_\mathrm{max})^{-1})$. The weight $w_\xi^{t-1}$ is reduced by the fraction obtained from the product of $d_\chi^{m,\mu\nu}$ and the absolute current weight update $|u_\xi^\tau|\cdot P_\xi^\tau\cdot s_\xi^\tau$. We note that the sign of the weight update---equal to the sign of $g_\xi^\tau$---is only encoded in $u_\xi^\tau$. The inverse weight decay hyperparameter is the maximal absolute weight value in well-behaving optimizations, i.e., $u_\xi^\tau\leq\pm1$, preventing strong increases or decreases of weights. To obtain the new weight $w_\xi^t$, the current weight update $u_\xi^\tau\cdot P_\xi^\tau\cdot s_\xi^\tau$ is subtracted from the previous weight $w_\xi^{t-1}$. Hence, the previous weight is changed in the opposite direction than the sign of $g_\xi^\tau$ with an individually adapted learning rate.

The score value,
\begin{align}
S_\xi^\tau=\begin{cases}\hspace{-0.1cm}\begin{array}{l}S_\xi^{\tau-1}+\left(t_\mathrm{hist}\right)^{-1}g_\xi^\tau\cdot u_\xi^\tau\cdot P_\xi^\tau\cdot s_\xi^\tau\end{array}&\hspace{-0.1cm}\mathrm{for}\ \tau\leq t_\mathrm{hist}\vspace{0.075cm}\\
\hspace{-0.1cm}\begin{array}{l}\left[1-\left(t_\mathrm{hist}\right)^{-1}\right]S_\xi^{\tau-1}\\
+\left(t_\mathrm{hist}\right)^{-1}g_\xi^\tau\cdot u_\xi^\tau\cdot P_\xi^\tau\cdot s_\xi^\tau\end{array}&\hspace{-0.1cm}\mathrm{otherwise}\end{cases}\ ,\label{eq:score}
\end{align}
accounts for weight specific contributions to previous loss function decreases. It is inspired from the synaptic intelligence method \cite{Zenke2017}. The loss function decrease is estimated by the product of the moving average of the gradient $g_\xi^\tau$ and the weight update $u_\xi^\tau\cdot P_\xi^\tau\cdot s_\xi^\tau$ for each step $\tau$. For infinitesimally small changes in the opposite direction of the gradient, the product of gradient and change equals the respective loss function decrease. For larger updates in optimizations, as in the present case, the loss function decrease is typically overestimated by this product, but still reasonable. In addition, the gradient will be noisy if it is calculated for a subsample of the entire training data set. The sign inversion of the update is omitted in the calculation of the score value. Consequently, a higher positive score value represents a larger loss function decrease. The score value is initialized as $S_\xi^0=0$. For each $\tau\leq t_\mathrm{hist}$, $g_\xi^\tau\cdot u_\xi^\tau\cdot P_\xi^\tau\cdot s_\xi^\tau$ is summed with equal contribution $(t_\mathrm{hist})^{-1}$ to build an initial history. Afterwards, $S_\xi^\tau$ is calculated as an exponential moving average with decay parameter $1-(t_\mathrm{hist})^{-1}$. The score value identifies the most important weights for the accurate prediction of previous training data. These weights can then be restricted by the plasticity factor (Equation (\ref{eq:plasticity})), to balance the stability-plasticity ratio.

As a variant, CoRe can also use the sign of the moving average of the gradient $g_\xi^\tau$ as update factor $u_\xi^\tau$, i.e., $u_\xi^\tau=\mathrm{sgn}(g_\xi^\tau)$. Furthermore, CoRe can be employed without plasticity factors, i.e., $n_\mathrm{frozen,\chi}^{m,\mu\nu}=0$. The Adam optimizer is a special case of CoRe for the hyperparameter settings $\beta_1^\mathrm{a}=\beta_1^\mathrm{b}$, $n_\mathrm{frozen,\chi}^{m,\mu\nu}=0$, $\eta_+,\eta_-=1$, and $d_\chi^{m,\mu\nu}=0$. RPROP without the backtracking weight step is another special case for the hyperparameter settings $\beta_1^\mathrm{a},\beta_1^\mathrm{b},\beta_2=0$, $n_\mathrm{frozen,\chi}^{m,\mu\nu}=0$, and $d_\chi^{m,\mu\nu}=0$.

General recommendations of the hyperparameter values can be provided for $\beta_2=0.999$, $\epsilon=10^{-8}$, $\eta_-=0.5$, $\eta_+=1.2$, $s_\mathrm{min}=10^{-6}$, and $s_\mathrm{max}=1$, which are mostly in agreement with the Adam optimizer and RPROP. Typically, $s_\xi^0=10^{-3}$ is a good choice for the initial learning rate. The weight decay $d_\chi^{m,\mu\nu}$ depends on the desired range of weight values. For example, weights associated with the output neuron in atomic neural networks should not be restricted to allow for an arbitrary atomic energy value, while weights connecting input and hidden layers or hidden and hidden layers applying hyperbolic tangents as activation functions can be restricted by using $d_\chi^{m,\mu\nu}=0.1$. $\beta_1^\mathrm{c}$ and $t_\mathrm{hist}$ depend on the convergence speed and the total number of epochs. For example, $\beta_1^\mathrm{c}=500$ and $t_\mathrm{hist}=500$ were employed in this work, while the total number of epochs were 1500, 2000, and 2500. In addition, $n_\mathrm{frozen,\chi}^{m,\mu\nu}$ was set to $1\%$ of the number of weights $a_{\kappa\lambda}^{m,\mu\nu}$ and $b_{\lambda}^{m,\nu}$ in their respective group, except for those weights associated with the output neuron. $\beta_1^\mathrm{a}$ and $\beta_1^\mathrm{b}$ need to be smaller for rapidly and strongly changing gradients. For example, $\beta_1^\mathrm{a}=0.45$ and $\beta_1^\mathrm{b}=0.7,0.725$ were used in this work. We note that for highly uncorrelated training data an increase of the subsample size can be beneficial to reduce gradient fluctuations.

%%%%%%%%%%%%%%%%%%%%%%%%%%%%%%%%%%%%%%%%%%%%%%%%%%%%%%%%%%%%%%%%%%%%%%%%%%%%%%%%%%%%%%%%%%%%%%%%%%%%%%%%
\subsection{Lifelong Adaptive Data Selection}
%%%%%%%%%%%%%%%%%%%%%%%%%%%%%%%%%%%%%%%%%%%%%%%%%%%%%%%%%%%%%%%%%%%%%%%%%%%%%%%%%%%%%%%%%%%%%%%%%%%%%%%%

Stochastic optimization, that is gradient calculation on a subsample of the training data in each epoch, can retain the computational demand on a manageable level for large data sets. However, random subsamples may lead to an inefficient approach because various data are typically represented at different levels of accuracy. This issue will be even more severe if additional training data are added during the training process in lifelong machine learning. Moreover, redundant and incorrect data need to be removed during the training process to be able to learn autonomously from a continuous stream of new data. Still, rehearsal of representative old data is very important to avoid catastrophic forgetting.

To find a solution for these issues, we developed an algorithm for lifelong adaptive data selection. The first key ingredient of the algorithm is an adaptive selection factor $S_\mathrm{hist}^r$ for each training conformation $r$ depending on the conformation's loss function contribution and the training history. The second key ingredient is a balancing scheme for the stochastic choice of good and bad represented training data. The algorithm intends to exclude redundant and incorrect data and to improve the training performance by balancing adjustment to new or bad represented data and rehearsal of good represented data.

\begin{table}[htb!]
\centering
\begin{tabular}{p{8cm}}
\hline\vspace{-0.175cm}
\textbf{Algorithm 1:} Choice of the training data subsample to be fitted in a training epoch in the lifelong adaptive data selection algorithm. All vector operations are element-wise. Assignment statements including conditions change only vector entries for which the condition is true.\vspace{0.075cm}\\
\hline\vspace{-0.175cm}
$N_\mathrm{fit}\leftarrow\mathrm{min}\left[N_\mathrm{fit},\mathrm{len}\left(\mathbf{S}_\mathrm{hist}^{>0}\right)\right]$\\
$N_\mathrm{good}\leftarrow \mathrm{int}\left(p_\mathrm{good}\cdot N_\mathrm{fit}\right)$\\
$N_\mathrm{bad}\leftarrow N_\mathrm{fit}-N_\mathrm{good}$\\
$L_\mathrm{old}^\mathrm{max}\leftarrow\mathrm{max}\left(\mathbf{L}_\mathrm{old}^{\backslash\mathrm{NaN}}\right)$\\
$\mathbf{P}_\mathrm{bad}\leftarrow \mathbf{S}_\mathrm{hist}\dfrac{\mathbf{L}_\mathrm{old}}{L_\mathrm{old}^\mathrm{max}}$\\
$\mathbf{P}_\mathrm{bad}\leftarrow\mathrm{max}\left(\mathbf{S}_\mathrm{hist}\right)\ \ \mathrm{if}\ P_\mathrm{bad}^r=\mathrm{NaN}$\\
$\mathbf{P}_\mathrm{bad}\leftarrow\dfrac{\mathbf{P}_\mathrm{bad}}{\mathrm{sum}\left(\mathbf{P}_\mathrm{bad}\right)}$\\
$\mathbf{D}_\mathrm{fit}\leftarrow\mathrm{random{\_}choice}\left(\mathbf{D},\ \mathbf{P}_\mathrm{bad},\ N_\mathrm{bad}\right)$\\
$\mathbf{P}_\mathrm{good}\leftarrow\mathbf{S}_\mathrm{hist}^{\backslash\mathrm{fit}}\left[\mathbf{1}-\dfrac{\mathbf{L}_\mathrm{old}^{\backslash\mathrm{fit}}}{L_\mathrm{old}^\mathrm{max}}\right]$\\
$P_\mathrm{good}^\mathrm{min}\leftarrow \mathrm{min}\left(\mathbf{P}_\mathrm{good}^{>0},1\right)\cdot\epsilon'$\\
$\mathbf{P}_\mathrm{good}\leftarrow P_\mathrm{good}^\mathrm{min}\ \ \mathrm{if}\ L_{\mathrm{old},r}^{\backslash\mathrm{fit}}=L_\mathrm{old}^\mathrm{max}$\\
$\mathbf{P}_\mathrm{good}\leftarrow P_\mathrm{good}^\mathrm{min}\ \ \mathrm{if}\ P_\mathrm{good}^r=\mathrm{NaN}$\\
$\mathbf{P}_\mathrm{good}\leftarrow\dfrac{\mathbf{P}_\mathrm{good}}{\mathrm{sum}\left(\mathbf{P}_\mathrm{good}\right)}$\\
$\mathbf{D}_\mathrm{fit}\leftarrow\mathbf{D}_\mathrm{fit}\cup\mathrm{random{\_}choice}\left(\mathbf{D}^{\backslash\mathrm{fit}},\ \mathbf{P}_\mathrm{good},\ N_\mathrm{good}\right)$\vspace{0.075cm}\\
\hline
\end{tabular}
\end{table}

The lifelong adaptive data selection algorithm can be subdivided into a part carried out before the loss function calculation and a part after this calculation. In the first part (Algorithm 1) the training data subsample is chosen for which the loss function and the respective gradients are calculated. This subsample is indicated by the index ``fit'' as these data are used for fitting the weights. The number of conformations to be fitted per epoch $N_\mathrm{fit}$ is a training hyperparameter and can be chosen based on the data set size and the correlation between the data points. Since conformations can be excluded by an adaptive selection factor $S_\mathrm{hist}^r$ (Algorithm 2), $N_\mathrm{fit}$ cannot be larger than the number of conformations with $S_\mathrm{hist}^r>0$, i.e., $\mathrm{len}\left(\mathbf{S}_\mathrm{hist}^{>0}\right)$. The training data subsample is split into a set of $N_\mathrm{good}$ already well represented conformations and $N_\mathrm{bad}$ new or insufficiently represented conformations. The fraction of good data is defined by $p_\mathrm{good}$ (Algorithm 2), while $N_\mathrm{good}$ has to be an integer. $p_\mathrm{good}$ is initialized as 0.

To determine the probabilities $\mathbf{P}_\mathrm{bad}$ and $\mathbf{P}_\mathrm{good}$ for conformations to be chosen as bad or good data, respectively, the contributions of the conformations to the loss function are employed. The last calculated contributions $\mathbf{L}_\mathrm{old}$ (or Not a Number (NaN) if no contribution was calculated for a conformation so far) are divided by the maximal, non-NaN contribution $\mathbf{L}_\mathrm{old}^\mathrm{max}$. These quotients are multiplied by the conformation specific adaptive selection factors $\mathbf{S}_\mathrm{hist}$ to yield the probabilities $\mathbf{P}_\mathrm{bad}$. $\mathbf{S}_\mathrm{hist}$ is initialized as unity vector. All NaN components of $\mathbf{P}_\mathrm{bad}$ are set to the maximum of $\mathbf{S}_\mathrm{hist}$. For normalization $\mathbf{P}_\mathrm{bad}$ is divided by the sum of its components. A random choice of $N_\mathrm{bad}$ conformations with probabilities $\mathbf{P}_\mathrm{bad}$ is selected from the training data set $\mathbf{D}$ to be part of the training data subsample.

To obtain $N_\mathrm{good}$ good data from the remaining data set $\mathbf{D}^{\backslash\mathrm{fit}}$, the probability $\mathbf{P}_\mathrm{good}$ is determined. $\mathbf{P}_\mathrm{good}$ is proportional to the difference between a unity vector $\mathbf{1}$ and the quotient of $\mathbf{L}_\mathrm{old}^{\backslash\mathrm{fit}}$ and $L_\mathrm{old}^\mathrm{max}$. The minimum non-zero probability of $\mathbf{P}_\mathrm{good}$ is determined and multiplied by $\epsilon'\gtrapprox0$ to obtain a very low probability $P_\mathrm{good}^\mathrm{min}$ for non-excluded but unfavored conformations. $\mathbf{P}_\mathrm{good}$ entries are set to $P_\mathrm{good}^\mathrm{min}$ if the loss function contribution of the corresponding conformation is equal to the maximal contribution or if the $\mathbf{P}_\mathrm{good}$ entry is NaN. Subsequently, $\mathbf{P}_\mathrm{good}$ is divided by the sum of its contributions. The good data subsample is a random choice of $N_\mathrm{good}$ data from $\mathbf{D}^{\backslash\mathrm{fit}}$ with probabilities $\mathbf{P}_\mathrm{good}$. The union of the selected bad and good data is then employed as training data subsample to calculate the loss function.

\begin{table}[htb!]
\centering
\begin{tabular}{p{8cm}}
\hline\vspace{-0.175cm}
\textbf{Algorithm 2:} Update of the adaptive selection factors for each conformation in the training data subsample in the lifelong adaptive data selection algorithm. All vector operations are element-wise. Assignment statements including conditions change only vector entries for which the condition is true.\vspace{0.075cm}\\
\hline\vspace{-0.175cm}
$\mathbf{L}_\mathrm{new}^\mathrm{fit}\leftarrow \dfrac{q^2\mathbf{L}_E^\mathrm{fit}+\tfrac{1}{3}\mathbf{L}_F^\mathrm{fit}}{\mathbf{N}_\mathrm{atom}^\mathrm{fit}}$\\
$\mathbf{L}_\mathrm{rel}\leftarrow\dfrac{\mathbf{L}_\mathrm{new}^\mathrm{fit}}{\overline{L}_\mathrm{new}^\mathrm{fit}}$\\
$\mathbf{X}\leftarrow X^r+1\ \ \mathrm{if}\ L_\mathrm{rel}^r>T_\mathbf{X}$\\
$\mathbf{X}\leftarrow 0\ \ \mathrm{if}\ L_\mathrm{rel}^r\leq T_\mathbf{X}$\\
$\mathbf{S}_\mathrm{hist}\leftarrow\mathrm{max}\left(1,S_\mathrm{hist}^r\right)\ \ \mathrm{if}\ L_\mathrm{rel}^r\geq T_F^1$\\
$\mathbf{S}_\mathrm{hist}\leftarrow\mathrm{min}\left(S_\mathrm{hist}^r,1\right)\ \ \mathrm{if}\ L_\mathrm{rel}^r\leq T_F^2$\\
$\mathbf{S}_\mathrm{hist}\leftarrow S_\mathrm{hist}^r\cdot F_{--}\ \ \mathrm{if}\ L_\mathrm{rel}^r<T_F^1\land L_\mathrm{new}^r\leq L_\mathrm{old}^r$\\
$\mathbf{S}_\mathrm{hist}\leftarrow S_\mathrm{hist}^r\cdot F_{-}\ \ \mathrm{if}\ L_\mathrm{rel}^r<T_F^1\land L_\mathrm{new}^r>L_\mathrm{old}^r$\\
$\mathbf{S}_\mathrm{hist}\leftarrow S_\mathrm{hist}^r\cdot F_+\ \ \mathrm{if}\ T_F^2<L_\mathrm{rel}^r\leq T_F^3\land L_\mathrm{new}^r>L_\mathrm{old}^r$\\
$\mathbf{S}_\mathrm{hist}\leftarrow S_\mathrm{hist}^r\cdot F_{++}\ \ \mathrm{if}\ L_\mathrm{rel}^r>T_F^3\land L_\mathrm{new}^r>L_\mathrm{old}^r$\\
$\mathbf{S}_\mathrm{hist}\leftarrow 0\ \ \mathrm{if}\ X^r\geq N_\mathbf{X}$\\
$\mathbf{S}_\mathrm{hist}\leftarrow 0\ \ \mathrm{if}\ S_\mathrm{hist}^r<S_\mathrm{hist}^\mathrm{min}\lor S_\mathrm{hist}^r>S_\mathrm{hist}^\mathrm{max}$\\
$p_\mathrm{good}\leftarrow\mathrm{clip}\left[p_\mathrm{good}+\mathrm{sgn}\left(\overline{L}_\mathrm{new}^\mathrm{fit}-\overline{L}_\mathrm{old}^\mathrm{fit}\right)p_\pm,\ 0,\ p_\mathrm{good}^\mathrm{max}\right]$\\
$\mathbf{L}_\mathrm{old}^\mathrm{fit}\leftarrow\mathbf{L}_\mathrm{new}^\mathrm{fit}$\\
$\overline{L}_\mathrm{old}^\mathrm{fit}\leftarrow\overline{L}_\mathrm{new}^\mathrm{fit}$\vspace{0.075cm}\\
\hline
\end{tabular}
\end{table}

To update the adaptive selection factors $\mathbf{S}_\mathrm{hist}$, the new loss function contributions $\mathbf{L}_\mathrm{new}^\mathrm{fit}$ of the conformations in the training data subsample are used (Algorithm 2). Therefore, a weighted sum of energy loss $\mathbf{L}_E^\mathrm{fit}$ and force loss $\mathbf{L}_F^\mathrm{fit}$ is employed, which is divided by the number of atoms of the respective conformations $\mathbf{N}_\mathrm{atom}^\mathrm{fit}$. $\mathbf{L}_\mathrm{rel}$, which is the quotient of $\mathbf{L}_\mathrm{new}^\mathrm{fit}$ and the total loss $\overline{L}_\mathrm{new}^\mathrm{fit}$ of the training epoch as defined in Equation (\ref{eq:loss}), is compared to different thresholds $T_F^1<1<T_F^2<T_F^3<T_\mathbf{X}$ to update $\mathbf{S}_\mathrm{hist}$ and the exclusion strike counter $\mathbf{X}$. The latter is introduced to exclude prediction outliers. Under the assumption that the model is reasonable, outliers are presumably incorrect data points. Therefore, $\mathbf{X}$---initialized as zero vector---counts how many consecutive evaluations $L_\mathrm{rel}^r$ is greater than the threshold $T_\mathbf{X}$. If $X^r$ reaches the hyperparameter $N_\mathbf{X}$, the adaptive selection factor $S_\mathrm{hist}^r$ is set to zero and the respective conformation is thus excluded from further training.

The adaptive selection factors $\mathbf{S}_\mathrm{hist}$ are further modified depending on the following conditions: If $L_\mathrm{rel}^r\geq T_F^1$ and $S_\mathrm{hist}^r<1$, $S_\mathrm{hist}^r$ is set to one. If $L_\mathrm{rel}^r\leq T_F^2$ and $S_\mathrm{hist}^r>1$, $S_\mathrm{hist}^r$ is also changed to one. In this way, $S_\mathrm{hist}^r$ of well represented data ($L_\mathrm{rel}^r<T_F^1$) can be lower than one leading to lowering of the probability to be chosen for the training data subsample. On the other side, $S_\mathrm{hist}^r$ of bad represented data ($L_\mathrm{rel}^r>T_F^2$) can be greater than one increasing the selection probability. For conformations with $T_F^1\leq L_\mathrm{rel}^r\leq T_F^2$, $S_\mathrm{hist}^r$ is resetted to one, because $L_\mathrm{rel}^r$ is around the average loss contribution. In addition, $\mathbf{S}_\mathrm{hist}$ can be modified by decrease factors $F_{--}$ and $F_{-}$ and increase factors $F_{+}$ and $F_{++}$. In this way, $S_\mathrm{hist}^r$ tracks the training history to assess the training importance of the associated conformation $r$. The decrease and increase factors are initialized as $F_{-(-)}=(S_\mathrm{min})^{(N_{F_{-(-)}})^{-1}}$ and $F_{+(+)}=(S_\mathrm{max})^{(N_{F_{+(+)}})^{-1}}$, respectively. Consequently, the hyperparameters $N_{F_{--}}$, $N_{F_-}$, $N_{F_+}$, and $N_{F_{++}}$ define how many consecutive applications of the associated decrease and increase factor lead to an adaptive selection factor below and above the hyperparameters $S_\mathrm{min}$ and $S_\mathrm{max}$, respectively. For well represented data $S_\mathrm{hist}^r$ is decreased by $F_{--}$ if the loss function contribution stayed constant or decreased compared to the previously calculated one. Otherwise $S_\mathrm{hist}^r$ of well represented data is less strongly decreased by $F_-$. For $T_F^2\leq L_\mathrm{rel}^r\leq T_F^3$ and $L_\mathrm{new}^r>L_\mathrm{old}^r$, $S_\mathrm{hist}^r$ is increased by $F_+$, while it is increased by the larger factor $F_{++}$ for $L_\mathrm{rel}^r\geq T_F^3$ and $L_\mathrm{new}^r>L_\mathrm{old}^r$.

Afterwards, all $S_\mathrm{hist}^r$ below $S_\mathrm{min}$ are set to zero for data set reduction because the associated conformations are steadily well represented by the model. Thus, these data are presumably redundant and can be excluded since rehearsal of other data and/or other learning strategies are sufficient to not forget these conformations. All $S_\mathrm{hist}^r$ above $S_\mathrm{max}$ are also set to zero because the model was not able for many optimization steps to represent these conformations accurately. Therefore, under the assumption that the model is reasonable, these conformations are not consistent to the major part of the other data and can be removed from the training data. In this way, the training performance improves for the other data.

For adaptive balancing of learning bad represented data and not forgetting good represented data, the fraction of good data $p_\mathrm{good}$ is modified according to the change of the total loss $\overline{L}_\mathrm{new}^\mathrm{fit}-\overline{L}_\mathrm{old}^\mathrm{fit}$. $p_\mathrm{good}$ is increased by $p_\pm$ if the loss change is positive, while it is decreased by $p_\pm$ for negative loss change. $\overline{L}_\mathrm{old}$ is initialized as infinity and $p_\pm$ is defined as $p_\mathrm{good}^\mathrm{max}\cdot (N_p)^{-1}$, with the maximal $p_\mathrm{good}$ value $p_\mathrm{good}^\mathrm{max}$ and the total number of possible $p_\mathrm{good}$ values $N_p$. Lastly, the new loss function contributions $\mathbf{L}_\mathrm{new}^\mathrm{fit}$ overwrite the old loss function contributions $\mathbf{L}_\mathrm{old}^\mathrm{fit}$ for the conformations of the training data subsample and the new total loss value $\overline{L}_\mathrm{new}^\mathrm{fit}$ replaces the old total loss value $\overline{L}_\mathrm{old}^\mathrm{fit}$.

%%%%%%%%%%%%%%%%%%%%%%%%%%%%%%%%%%%%%%%%%%%%%%%%%%%%%%%%%%%%%%%%%%%%%%%%%%%%%%%%%%%%%%%%%%%%%%%%%%%%%%%%
\subsection{Uncertainty Quantification}\label{sec:uncertainty_quantification}
%%%%%%%%%%%%%%%%%%%%%%%%%%%%%%%%%%%%%%%%%%%%%%%%%%%%%%%%%%%%%%%%%%%%%%%%%%%%%%%%%%%%%%%%%%%%%%%%%%%%%%%%

An lMLP requires to predict the uncertainties of the energy $\Delta E$ and the atomic force components $\Delta F_{\alpha,n}$ in addition to their values for a chemical structure. The reason is that a prerequisite of an lMLP is that it can produce results at every training stage. Since the predictions at different training stages can vary, the lMLP needs to provide a measure of the prediction accuracy. Then, the predictions can be compared within their accuracy in which they should agree. In this way, a reliable method is obtained which still can adapt to new information over time. We note that uncertainty quantification also significantly advances other computational chemistry methods because overinterpretation of features beyond the method's accuracy can be prevented.

The uncertainty of machine learning models can be separated into contributions from noise in the reference data, bias of the model, and model variance contributions \cite{Heid2023}. The handling of noisy training data is addressed by the lifelong adaptive data selection, while noise of the test data does not influence the model performance but only the performance evaluation. For correctly converged electronic structure calculations the noise is determined by the convergence thresholds and thus small for MLP reference data. Model bias is caused by limitations in the model architecture, representation by the descriptors, and reference data set size. Therefore, the model architecture needs to be optimized for different applications and elaborate descriptors need to be tested to keep this error contribution small. Lifelong learning addresses the remaining part of incomplete reference data. Model variance error can be reliably reduced by ensemble (or committee) models and its size can be predicted by statistics. The respective uncertainty estimation is based on the huge flexibility of the mathematical expressions underlying machine learning models like artificial neural networks. For well trained conformation space the ensemble member predictions presumably agree, while the predictions are arbitrary for unknown conformation space. Therefore, they can spread largely for an ensemble of independently and differently trained MLPs, i.e., using, for example, different initial weights.

Consequently, ensembles of MLPs have been used for a straightforward uncertainty quantification of model variance contributions \cite{Peterson2017, Smith2018, Smith2019, Devereux2020, Musil2019, Imbalzano2021}. The final energy prediction $\overline{E}$ is obtained by the mean of the ensemble of $N_\mathrm{MLP}$ MLP energies $E_p$,
\begin{align}
\overline{E}=\frac{1}{N_\mathrm{MLP}}\sum_{p=1}^{N_\mathrm{MLP}}E_p\ .
\end{align}
Analogously, the atomic force components $\overline{F}_{\alpha,n}$ are calculated as mean of the negative energy gradient with respect to the Cartesian coordinates $\alpha_n$ of an MLP ensemble,
\begin{align}
\overline{F}_{\alpha,n}=-\frac{1}{N_\mathrm{MLP}}\sum_{p=1}^{N_\mathrm{MLP}}\frac{\partial E_p}{\partial \alpha_n}\ .
\end{align}
The energy uncertainties $\Delta\overline{E}$ can then be obtained from the sample standard deviation of the ensemble member energy predictions,
\begin{align}
\begin{split}
&\Delta\overline{E}=\mathrm{max}\left\{\mathrm{RMSE}(\overline{E}),\right.\\
&\left.\left[\frac{c^2}{N_\mathrm{MLP}-1}\sum_{p=1}^{N_\mathrm{MLP}}\left(E_p-\overline{E}\right)^2\right]^{\tfrac{1}{2}}\right\}\ ,
\end{split}
\end{align}
while the minimal uncertainty is defined by the root mean square error $\mathrm{RMSE}(\overline{E})$ of the reference data. The atomic force component uncertainties $\Delta\overline{F}_{\alpha,n}$ can be analogously calculated.

The scaling factor $c$ is introduced to adjust the sample standard deviation providing a certain confidence interval that the uncertainty quantification is equal to or larger than the actual error with respect to the reference data. However, this uncertainty quantification misses model bias contributions and hence underestimates the error in cases where the lack of training data is the primary error source. Despite that, as also the model variance increases in these cases, this uncertainty quantification can still signal whether the error will be high even if the predicted uncertainty value is not accurate. For applications of MLPs this information is sufficient because small uncertainties can be well predicted and the occurrence of large errors requires in any way an improvement of the calculation which can be achieved by lifelong machine learning.

%%%%%%%%%%%%%%%%%%%%%%%%%%%%%%%%%%%%%%%%%%%%%%%%%%%%%%%%%%%%%%%%%%%%%%%%%%%%%%%%%%%%%%%%%%%%%%%%%%%%%%%%
\subsection{Perspective of Lifelong Machine Learning Potentials}
%%%%%%%%%%%%%%%%%%%%%%%%%%%%%%%%%%%%%%%%%%%%%%%%%%%%%%%%%%%%%%%%%%%%%%%%%%%%%%%%%%%%%%%%%%%%%%%%%%%%%%%%

On the long-term perspective an lMLP which learns more and more systems over time while the training benefits from prior learned knowledge of chemical interactions is the grand goal. The lMLP learns then in the same way as chemistry evolves, i.e., by systematically expanding prior knowledge. As this work is a proof of concept, the presented lifelong training strategies are naturally not in such an advanced state yet to reach this grand goal. In the short-term perspective with the presented algorithms, lMLPs will presumably be trained on data of related systems because the benefit is the largest and the difficulty is the lowest in this case. When the lifelong training strategies as well as the MLP models are developed to a more elaborate state, more and more information can be combined in one lMLP.

This work presents a comprehensive basis of ingredients for lMLPs. Future works need, for example, to develop algorithms for the continual expansion of the model architectures to avoid information capacity limitations of the underlying deep learning representations. For instance, neural network growth can be established by using differently sized architectures of the lMLP ensemble members. By tracking their relative performance the requirement to grow can be identified and individual ensemble members can be adapted. Moreover, coarse-grained structural identifiers like the bin and hash method \cite{Paleico2021} can be applied to augment the selection and exclusion criteria of training data used for rehearsal. In addition, they can be employed for fast checks whether a certain conformation is represented by similar ones in the training data. From this information unknown and badly represented conformations can be distinguished to avoid redundant reference calculations.

Current MLP models are often limited to the representation of a single electronic state. Thus, for example, only a specific total charge and spin multiplicity can often be represented by a single MLP. However, recent works target these limitations for more general MLP models \cite{Ko2021, Eckhoff2021, Westermayr2021} which can then be used as base model for lMLPs. A current restriction of MLPs is the requirement of consistent training data, i.e., all training data have to be calculated by the same method, basis set, etc. Therefore, MLP models need to be developed which, for example, treat different reference methods in a similar fashion as different electronic states to establish lMLPs learning from data of different reference methods.

Further, open, free, and accessible reference data of published work is obviously highly important. To continue the training of previous work, not only the final weight values but also the associated CoRe optimizer properties of the weights $\tau$, $g_\xi^\tau$, $h_\xi^\tau$, $s_\xi^\tau$, and $S_\xi^\tau$ and the lifelong adaptive data selection properties $L_\mathrm{old}^r$, $S_\mathrm{hist}^r$, and $X^r$ of every still required training conformation as well as the values $\overline{L}_\mathrm{old}$ and $p_\mathrm{good}$ need to be saved and made available. Moreover, autonomous workflows need to be set up in future work for a user-friendly development of an lMLP with direct interfaces to electronic structure and atomistic simulation software.

%%%%%%%%%%%%%%%%%%%%%%%%%%%%%%%%%%%%%%%%%%%%%%%%%%%%%%%%%%%%%%%%%%%%%%%%%%%%%%%%%%%%%%%%%%%%%%%%%%%%%%%%
\section{Computational Details}\label{sec:Comp_Details}
%%%%%%%%%%%%%%%%%%%%%%%%%%%%%%%%%%%%%%%%%%%%%%%%%%%%%%%%%%%%%%%%%%%%%%%%%%%%%%%%%%%%%%%%%%%%%%%%%%%%%%%%

In the HDNNP models of this work, each atomic neural network consisted of an input layer with $n_G=153$ (eeACSFs) or $156$ (ACSFs) neurons, three hidden layers with $n_1=102$, $n_2=61$, and $n_3=44$ neurons, and a single output neuron. The weight initialization is described in Section S1.1 of the Supporting Information. In predictions, the ensemble size $N_\mathrm{MLP}=10$ was applied with the error scaling factor $c=2$. We note that the ensemble size was not optimized for individual cases, but it can be in future applications depending on the improvement of value and uncertainty prediction versus additional computational demand.

For the training of each HDNNP, we split the reference data set randomly into $90\%$ training conformations and $10\%$ test conformations. In each training epoch, $10\%$ of all training conformations were used for fitting reference data sets A and B (Table \ref{tab:reference_data}) and $4.07\%$ were employed for fitting reference data set C. The latter yields the same number of trained conformations per epoch for reference data set C compared to B. This counting includes conformations with an adaptive selection factor of $S_\mathrm{hist}^r=0$ and conformations which were first added at a late training epoch. To represent the molecular structures of the reference conformations, eeACSF vectors were equipped with the parameters given in Table \ref{tab:eeACSF_parameters}. A larger than usual cutoff radius of $R_\mathrm{c}=12\,\text{\AA}$ was applied to avoid issues with electrostatic interactions, which are not in the focus of this work. However, future studies can easily adopt more elaborate schemes for electrostatics like fourth-generation HDNNPs \cite{Ko2021}.

\begin{table}[htb!]
\caption{All combinations of the listed parameters were applied for radial and angular eeACSFs, respectively. The cutoff radius was set to $R_\mathrm{c}=12\,\text{\AA}$ for all eeACSFs. $\gamma=\pm1$ was used except for $H^\mathrm{ang}=1$, where only $\gamma=1$ was applied.}
\begin{center}
\begin{tabular}{ll}
\hline\vspace{-0.325cm}\\
\multicolumn{2}{l}{Radial eeACSFs}\vspace{0.075cm}\\
\hline\vspace{-0.325cm}\\
$H^\mathrm{rad}$ & $1$, $n$, $m$, $\overline{n}$, $\overline{m}$\\
$\eta^\mathrm{rad}\,/\,\text{\AA}^{-2}$ & $0$, $0.010702$, $0.023348$, $0.044203$,\\
 & $0.066118$, $0.104168$, $0.180285$,\\
 & $0.370959$, $1.115414$\vspace{0.075cm}\\
\hline\vspace{-0.325cm}\\
\multicolumn{2}{l}{Angular eeACSFs}\vspace{0.075cm}\\
\hline\vspace{-0.325cm}\\
$H^\mathrm{ang}$ & $1$, $n$, $m$, $\overline{n}$, $\overline{m}$\\
$\eta^\mathrm{ang}\,/\,\text{\AA}^{-2}$ & $0.011238$, $0.090144$\\
$\lambda$ & $-1$, $1$\\
$\xi$ & $1$, $2.409421$, $9.996864$\vspace{0.075cm}\\
\hline
\end{tabular}
\end{center}
\label{tab:eeACSF_parameters}
\end{table}

The supervised training was performed on the total DFT energy minus the sum of the atomic DFT energies of the neutral free atoms in their lowest spin state (see Table S1 in the Supporting Information). All DFT data of S$_\text{N}$2 reactions were obtained for a total charge of $-1\,e$, with $e$ being the elementary charge, and a spin multiplicity of $1$. The DFT calculations were performed with the quantum chemistry program ORCA (version 5.0.3) \cite{Neese2012, Neese2022}. The PBE exchange-correlation functional \cite{Perdew1996} was chosen with the def2-TZVP basis set \cite{Weigend2005} and use of the RI-J approximation. In addition, D3 semi-classical dispersion corrections \cite{Grimme2010} with Becke-Johnson damping \cite{Johnson2005, Grimme2011} were applied. To balance the training on the reference DFT energies and forces, the loss function hyperparameter was set to $q=10.9$.

For SGD optimizations the best found learning rate of 0.00075 was applied. RPROP optimizations were performed with the hyperparameters recommended in Reference \citenum{Riedmiller1993}. An exception was the initial learning rate, for which 0.001 was found to be the optimal choice in this work. For the Adam optimizer the same hyperparameters were used as in Reference \citenum{Kingma2015} since no general improvement by hyperparameter variations was found.

The hyperparameters of the CoRe optimizer were: $\beta_1^\mathrm{a}=0.45$, $\beta_1^\mathrm{b}=0.7,0.725$, $\beta_1^\mathrm{c}=500$, $\beta_2=0.999$, $\epsilon=10^{-8}$, $\eta_-=0.5$, $\eta_+=1.2$, $s_\mathrm{min}=10^{-6}$, $s_\mathrm{max}=1$, $s_\xi^0=10^{-3}$, and $t_\mathrm{hist}=500$. $n_\mathrm{frozen,\chi}^{m,\mu\nu}$ was set to $1\%$ of the number of weights $a_{\kappa\lambda}^{m,\mu\nu}$ and $b_{\lambda}^{m,\nu}$ in their respective group. Exceptions were those weights associated with the output neuron, for which $n_\mathrm{frozen,\chi}^{m,\mu\nu_\mathrm{max}}$ was 0. For the weights $\alpha$ and $\beta$, $n_\mathrm{frozen,\chi}^{m,\mu\nu}=0$ was applied. The weight decay hyperparameter $d_\chi^{m,\mu\nu}$ was $0.01$ for $\chi=\alpha,\beta$, 0 for $\nu=\nu_\mathrm{max}$, and $0.1$ otherwise. Training was performed for 1500, 2000, and 2500 epochs.

The hyperparameters of the lifelong adaptive data selection were: $S_\mathrm{hist}^\mathrm{min}=0.1$, $S_\mathrm{hist}^\mathrm{max}=100$, $\sqrt{T_F^1}=0.9$, $\sqrt{T_F^2}=1.2$, $\sqrt{T_F^3}=2.0$, $N_{F_{--}}=30$, $N_{F_-}=100$, $N_{F_+}=500$, $N_{F_{++}}=150$, $\sqrt{T_\mathbf{X}}=7.5$, $N_\mathbf{X}=5$, $p_\mathrm{good}^\mathrm{max}=\tfrac{2}{3}$, $N_p=20$, and $\epsilon'=10^{-6}$.

The lMLP software was written in Python and exploits the scientific computing package NumPy (version 1.21.6) \cite{Harris2020} and the machine learning framework PyTorch (version 1.12.1) \cite{Paszke2019}. It is available on Zenodo (DOI: 10.5281/zenodo.7912832) alongside the reference data sets, generated output of this work, and scripts to analyze and plot this output.

For a fair and reliable comparison of the different MLP trainings, we performed for each setting 20 trainings with different random numbers used in the splitting of training and test data, weight initialization, and training data selection process. For each setting, the best ten MLPs were employed in the analysis to reduce the dependence on the explicit random numbers. In this way, the effect of outliers was minimized that can originate from an unfavorable initial parameter choice, which can affect all optimizers. Thus, the provided means and standard deviations represent the performance for well working cases.

As we employed different splittings of training and test data in the training of ensemble members in this work, the ensemble prediction of the reference data mixes training and test data predictions. While this approach efficiently uses all reference data for training, unbiased validation of the ensemble needs to be performed on additional data.

%%%%%%%%%%%%%%%%%%%%%%%%%%%%%%%%%%%%%%%%%%%%%%%%%%%%%%%%%%%%%%%%%%%%%%%%%%%%%%%%%%%%%%%%%%%%%%%%%%%%%%%%
\section{Results and Discussion}\label{sec:Results}
%%%%%%%%%%%%%%%%%%%%%%%%%%%%%%%%%%%%%%%%%%%%%%%%%%%%%%%%%%%%%%%%%%%%%%%%%%%%%%%%%%%%%%%%%%%%%%%%%%%%%%%%

%%%%%%%%%%%%%%%%%%%%%%%%%%%%%%%%%%%%%%%%%%%%%%%%%%%%%%%%%%%%%%%%%%%%%%%%%%%%%%%%%%%%%%%%%%%%%%%%%%%%%%%%
\subsection{Reference Data}
%%%%%%%%%%%%%%%%%%%%%%%%%%%%%%%%%%%%%%%%%%%%%%%%%%%%%%%%%%%%%%%%%%%%%%%%%%%%%%%%%%%%%%%%%%%%%%%%%%%%%%%%

\begin{figure}[htb!]
\centering
\includegraphics[width=\columnwidth]{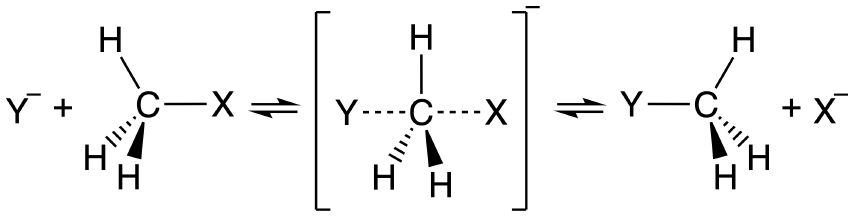}
\caption{S$_\text{N}$2 reaction at a methyl carbon atom with leaving group X and nucleophile Y.}\label{fig:SN2_reaction}
\end{figure}

For the comparison of the performance of ACSFs and eeACSFs, a reference data set A was constructed containing ten different gas-phase S$_\text{N}$2 reactions. These reactions were represented by 2026 reference conformations and their respective energies and atomic force components obtained from the reference method. The reference conformations were different chemical structures which sampled the conformation space of the S$_\text{N}$2 reactions including the leaving groups X$^-$\,=\,Cl$^-$, I$^-$ and nucleophiles Y$^-$\,=\,Cl$^-$, HCC$^-$, I$^-$ for central methyl carbon atoms and tertiary \textit{tert}-butyl carbon atoms (Figure \ref{fig:SN2_reaction}). We note that steric hindrance leads to high energy barriers in S$_\text{N}$2 reactions of tertiary carbon atoms. This set of S$_\text{N}$2 reactions included only four different elements so that the application of ACSFs was still feasible in terms of computational demand. The combinatorial growth of the ACSF vector with the number of elements hampers the application to reference data including more elements.

By contrast, the size of the eeACSF vector stays constant with the number of elements enabling training of reference data containing an arbitrary number of elements. To test eeACSFs, a reference data set B was compiled consisting of 8600 conformations including 42 different S$_\text{N}$2 reactions with leaving groups X$^-$\,=\,Cl$^-$, I$^-$, nucleophiles Y$^-$\,=\,Br$^-$, Cl$^-$, F$^-$, H$_2$N$^-$, H$_3$CO$^-$, HCC$^-$, HO$^-$, HS$^-$, HSe$^-$, I$^-$, NC$^-$, and central methyl and \textit{tert}-butyl carbon atoms.

Both reference data sets A and B contained conformations obtained in constrained DFT optimizations. The distances carbon-leaving group and nucleophile-carbon were set in the range from $1.05$ to $5.25\,\text{\AA}$ by an irregular sampling approach. A preliminary grid of distance values, which was more dense for smaller distances, was defined with individual minimal distances for each S$_\text{N}$2 reaction system. To obtain the final distance values, random changes on the grid values were applied with the constraint that the values cannot get smaller or larger than adjacent grid values..

To add the representation of structural distortions, 12551 conformations of the S$_\text{N}$2 reaction systems with central methyl carbon atoms were generated by random atomic displacements. For each conformation obtained in constrained DFT optimizations, three new conformations were generated by randomly displacing all atomic positions inside atom-centered spheres with the same radii of $0.05$, $0.1$, and $0.15\,\text{\AA}$. If some interatomic distances turned out to be too small, the process was restarted with another set of random displacements. These conformations together with those of reference data set B formed reference data set C. We emphasize that these reference data were only employed in performance evaluations of lifelong learning and can be insufficient to carry out atomistic simulations because relevant reference conformations can be missing.

The only preprocessing of the reference data sets was a restriction to maximal absolute atomic force components of $15\,\mathrm{eV}\,\text{\AA}^{-1}$ to exclude those conformations which only occur under extreme conditions. The energy and atomic force component ranges and standard deviations of the different data sets are compiled in Table \ref{tab:reference_data} and are referred to in performance comparisons in the following sections. The mean energy ranges and standard deviations for the individual S$_\text{N}$2 reaction systems are provided in Table S4 in the Supporting Information.

\begin{table}[htb!]
\caption{Number of S$_\text{N}$2 reactions $N_\mathrm{S_\mathrm{N}2}$, conformations $N_\mathrm{conf}$, atoms in total $N_\mathrm{atom}^\mathrm{total}$, and elements $N_\mathrm{elem}$ for the different reference data sets A, B, and C. Additionally, the energy range $E^\mathrm{ref}_\mathrm{range}$ and standard deviation $E^\mathrm{ref}_\mathrm{std}$ and the atomic force component range $F_{\alpha,n,\mathrm{range}}^\mathrm{ref}$ and standard deviation $F_{\alpha,n,\mathrm{std}}^\mathrm{ref}$ are provided.}
\begin{center}
\begin{tabular}{llll}
\hline\vspace{-0.325cm}\\
Reference data set & A & B & C\vspace{0.075cm}\\
\hline\vspace{-0.325cm}\\
$N_\mathrm{S_\mathrm{N}2}$ & 10 & 42 & 42\\
$N_\mathrm{conf}$ & 2026 & 8600 & 21151\\
$N_\mathrm{atom}^\mathrm{total}$ & 22983 & 100117 & 190065\\
$N_\mathrm{elem}$ & 4 & 10 & 10\\
$E^\mathrm{ref}_\mathrm{range}\,/\,\mathrm{meV\,atom}^{-1}$ & 1715.3 & 1855.0 & 2018.0\\
$E^\mathrm{ref}_\mathrm{std}\,/\,\mathrm{meV\,atom}^{-1}$ & 382.6 & 373.8 & 391.5\\
$F_{\alpha,n,\mathrm{range}}^\mathrm{ref}\,/\,\mathrm{meV}\,\text{\AA}^{-1}$ & 29883 & 29984 & 29998\\
$F_{\alpha,n,\mathrm{std}}^\mathrm{ref}\,/\,\mathrm{meV}\,\text{\AA}^{-1}$ & 1098 & 1066 & 1706\vspace{0.075cm}\\
\hline
\end{tabular}
\end{center}
\label{tab:reference_data}
\end{table}

%%%%%%%%%%%%%%%%%%%%%%%%%%%%%%%%%%%%%%%%%%%%%%%%%%%%%%%%%%%%%%%%%%%%%%%%%%%%%%%%%%%%%%%%%%%%%%%%%%%%%%%%
\subsection{Element-Embracing Atom-Centered Symmetry Functions}
%%%%%%%%%%%%%%%%%%%%%%%%%%%%%%%%%%%%%%%%%%%%%%%%%%%%%%%%%%%%%%%%%%%%%%%%%%%%%%%%%%%%%%%%%%%%%%%%%%%%%%%%

For four elements the sizes of the optimized ACSF vector (156) and eeACSF vector (153) are similar (see Table S2 in the Supporting Information for parameters of the ACSFs). Therefore, this number of elements is the turning point at which the eeACSF representation becomes computationally advantageous. Table \ref{tab:descriptor_accuracy} shows that the representation by ACSFs and eeACSFs of reference data set A yields HDNNPs with similar RMSEs for the test data justifying the structural representation by eeACSFs (see Figures S3 (a) and (b) in the Supporting Information for the prediction error distribution of the ensemble). The accuracy of HDNNPs using ACSFs is on average slightly better. This trend is expected due to the full separation of contributions from different element combinations, while these contributions are mixed in eeACSFs. However, the eeACSF representation is less prone to overfitting according to the given training and test RMSEs. The reason may be that every eeACSF value depends on all neighbor atoms inside the cutoff radius, while an ACSF value depends only on certain neighbor atoms. The latter may adjust better to very specific environments but worsens generalization and transferability. The Figures S4 (a) and (b) and S5 in the Supporting Information show that the convergence and training process is similar for ACSF and eeACSF representations.

\begin{table}[htb!]
\caption{RMSE values of individual HDNNPs, i.e., before ensembling, and the ensemble trained on reference data set A using ACSF or eeACSFs. The CoRe optimizer and lifelong adaptive data selection were applied for 2000 epochs.}
\begin{center}
\begin{tabular}{lll}
\hline\vspace{-0.325cm}\\
Individual HDNNPs & ACSF & eeACSF\vspace{0.075cm}\\
\hline\vspace{-0.325cm}\\
$\mathrm{RMSE}(E^\mathrm{train})\,/\,\mathrm{meV\,atom}^{-1}$ & $2.0\pm0.2$ & $2.5\pm0.3$\\
$\mathrm{RMSE}(E^\mathrm{test})\,/\,\mathrm{meV\,atom}^{-1}$ & $2.3\pm0.2$ & $2.8\pm0.3$\\
$\mathrm{RMSE}(F_{\alpha,n}^\mathrm{train})\,/\,\mathrm{meV}\,\text{\AA}^{-1}$ & $59\pm3$ & $73\pm3$\\
$\mathrm{RMSE}(F_{\alpha,n}^\mathrm{test})\,/\,\mathrm{meV}\,\text{\AA}^{-1}$ & $87\pm9$ & $92\pm8$\vspace{0.075cm}\\
\hline\vspace{-0.325cm}\\
Ensemble\vspace{0.075cm}\\
\hline\vspace{-0.325cm}\\
$\mathrm{RMSE}(\overline{E})\,/\,\mathrm{meV\,atom}^{-1}$ & 0.9 & 1.3\\
$\mathrm{RMSE}(\overline{F}_{\alpha,n})\,/\,\mathrm{meV}\,\text{\AA}^{-1}$ & 34 & 44\vspace{0.075cm}\\
\hline
\end{tabular}
\end{center}
\label{tab:descriptor_accuracy}
\end{table}

To train the ten element containing reference data set B, the size of the ACSF vector would be 750 to obtain the same resolution by the parameters $\eta^\mathrm{rad}$, $\eta^\mathrm{ang}$, $\lambda$, and $\xi$. The resulting high computational demand can be prevented by applying eeACSFs with a constant vector size of 153. The accuracy of individual HDNNPs is $\mathrm{RMSE}(E^\mathrm{train})=(3.9\pm0.4)\,\mathrm{meV\,atom}^{-1}$, $\mathrm{RMSE}(E^\mathrm{test})=(4.5\pm0.6)\,\mathrm{meV\,atom}^{-1}$, $\mathrm{RMSE}(F_{\alpha,n}^\mathrm{train})=(99\pm7)\,\mathrm{meV}\,\text{\AA}^{-1}$, and $\mathrm{RMSE}(F_{\alpha,n}^\mathrm{test})=(116\pm4)\,\mathrm{meV}\,\text{\AA}^{-1}$. The higher accuracy of the results in Table \ref{tab:descriptor_accuracy} is a reason of the fewer and less complex reference data to be trained, while the model architecture and training hyperparameters remained unchanged. However, especially the HDNNP ensemble accuracy of $\mathrm{RMSE}(\overline{E})=2.6\,\mathrm{meV\,atom}^{-1}$ and $\mathrm{RMSE}(\overline{F}_{\alpha,n})=64\,\mathrm{meV}\,\text{\AA}^{-1}$ (see Figures S6 (a) and (b) in the Supporting Information for the prediction error distribution) is comparable to other state-of-the-art MLPs trained for less elements \cite{Behler2021} and evidences that eeACSFs are able to represent the different local atomic environments including various neighbor elements. Further, the significantly improved accuracy of the HDNNP ensemble compared to that of individual HDNNPs supports the use of an ensemble beyond the access of uncertainty quantification. For performance comparisons of HDNNPs with other MLPs we refer to the References \citenum{Zuo2020} and \citenum{Chen2022}.

The energy RMSE is slightly larger than usual due to the relatively broad energy range to be trained (Table \ref{tab:reference_data}). Moreover, the mean energy range for the individual S$_\text{N}$2 reaction systems is also broad with $747\,\mathrm{meV\,atom}^{-1}$ for those with central methyl carbon atoms and $438\,\mathrm{meV\,atom}^{-1}$ for those with central \textit{tert}-butyl carbon atoms. The atomic force component RMSE is somewhat lower than usual despite the broad range because a significant fraction of forces is close to zero.

%%%%%%%%%%%%%%%%%%%%%%%%%%%%%%%%%%%%%%%%%%%%%%%%%%%%%%%%%%%%%%%%%%%%%%%%%%%%%%%%%%%%%%%%%%%%%%%%%%%%%%%%
\subsection{Continual Resilient (CoRe) Optimizer}
%%%%%%%%%%%%%%%%%%%%%%%%%%%%%%%%%%%%%%%%%%%%%%%%%%%%%%%%%%%%%%%%%%%%%%%%%%%%%%%%%%%%%%%%%%%%%%%%%%%%%%%%

\begin{figure}[htb!]
\centering
\includegraphics[width=\columnwidth]{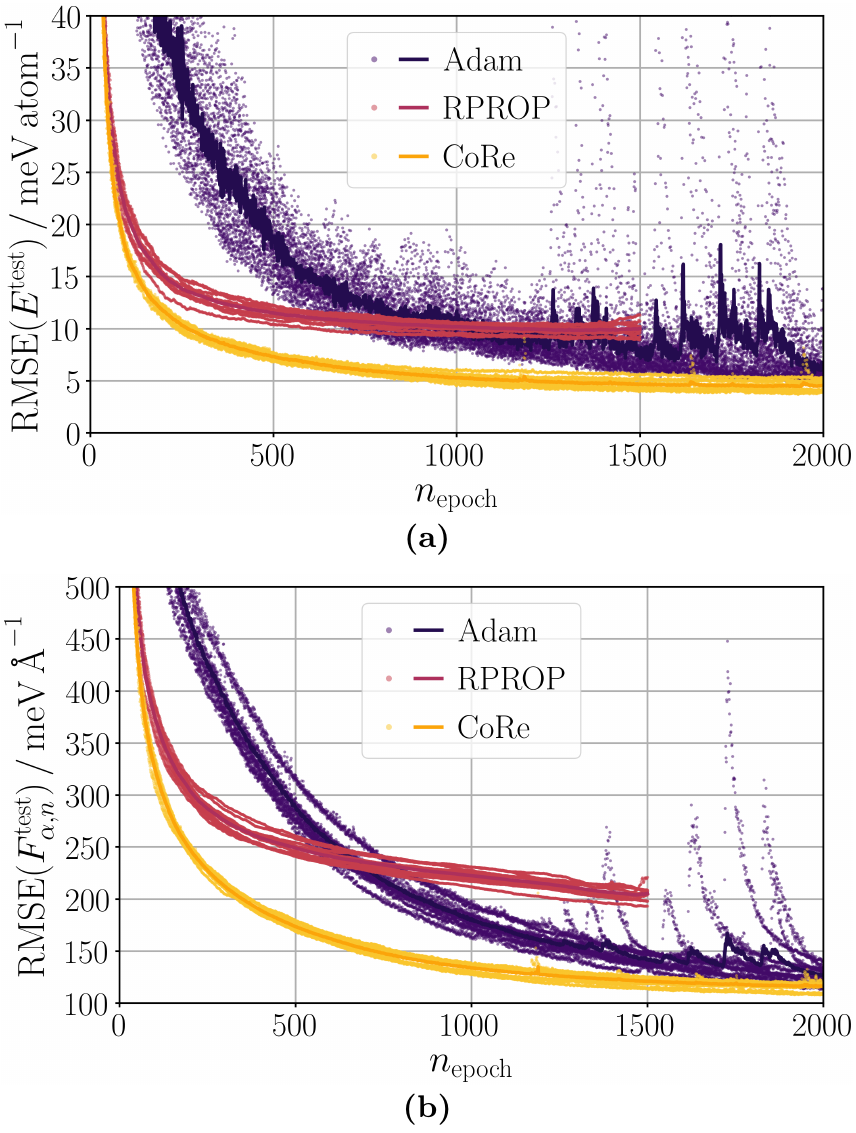}
\caption{Convergence of the optimizers Adam, RPROP, and CoRe for training reference data set B. The test set RMSE values of \textbf{(a)} energies $E^\mathrm{test}$ and \textbf{(b)} atomic force components $F_{\alpha,n}^\mathrm{test}$ are shown as a function of the training epoch $n_\mathrm{epoch}$. RMSE values of individual HDNNPs are represented by dots, while their mean, which is unequal to the ensemble RMSE value, is shown by a solid line. Lifelong adaptive data selection was applied in the optimizations.}\label{fig:optimizer_E_F}
\end{figure}

Accurate MLPs can only be obtained if the optimizer can efficiently and reliably find apposite weight values in the high-dimensional parameter space. Training with a fixed learning rate as in SGD yields HDNNPs of poor accuracy. Most RMSE values of SGD results for reference data set B listed in Table \ref{tab:optimizer_accuracy} are almost an order of magnitude larger than those of CoRe results highlighting the importance of the optimizer. We found RPROP and the Adam optimizer to be the best performing optimizers available in PyTorch 1.12.1 for the given machine learning model and reference data. We note that RPROP is intended for batch learning on all data at once, while we perform stochastic optimization with RPROP. RPROP converges fast and smooth, but plateaus at not satisfying accuracy (Figures \ref{fig:optimizer_E_F} (a) and (b)). By contrast, the Adam optimizer requires more steps to reach the accuracy of RPROP, but it is able to reach lower RMSE values in the end. However, the convergence is noisy and therefore hampers continual applications.

\begin{table*}[htb!]
\caption{RMSE values of individual HDNNPs and the ensemble trained on reference data set B using the optimizers SGD, RPROP, Adam, and CoRe. The optimizers and lifelong adaptive data selection were applied for 1500 (RPROP) and 2000 (SGD, Adam, CoRe) epochs.}
\begin{center}
\begin{tabular}{lllll}
\hline\vspace{-0.325cm}\\
Individual HDNNPs & SGD & RPROP & Adam & CoRe\vspace{0.075cm}\\
\hline\vspace{-0.325cm}\\
$\mathrm{RMSE}(E^\mathrm{train})\,/\,\mathrm{meV\,atom}^{-1}$ & $37\pm7$ & $9.5\pm0.7$ & $6.1\pm1.5$ & $3.9\pm0.4$\\
$\mathrm{RMSE}(E^\mathrm{test})\,/\,\mathrm{meV\,atom}^{-1}$ & $37\pm7$ & $10.0\pm0.7$ & $6.4\pm1.5$ & $4.5\pm0.6$\\
$\mathrm{RMSE}(F_{\alpha,n}^\mathrm{train})\,/\,\mathrm{meV}\,\text{\AA}^{-1}$ & $564\pm17$ & $191\pm5$ & $119\pm8$ & $99\pm7$\\
$\mathrm{RMSE}(F_{\alpha,n}^\mathrm{test})\,/\,\mathrm{meV}\,\text{\AA}^{-1}$ & $557\pm11$ & $205\pm8$ & $127\pm8$ & $116\pm4$\vspace{0.075cm}\\
\hline\vspace{-0.325cm}\\
Ensemble\vspace{0.075cm}\\
\hline\vspace{-0.325cm}\\
$\mathrm{RMSE}(\overline{E})\,/\,\mathrm{meV\,atom}^{-1}$ & $33$ & $6.8$ & $3.9$ & $2.6$\\
$\mathrm{RMSE}(\overline{F}_{\alpha,n})\,/\,\mathrm{meV}\,\text{\AA}^{-1}$ & $529$ & $131$ & $95$ & $64$\vspace{0.075cm}\\
\hline
\end{tabular}
\end{center}
\label{tab:optimizer_accuracy}
\end{table*}

Our CoRe optimizer converges even faster than RPROP and reaches a better final accuracy than the Adam optimizer (Figures \ref{fig:optimizer_E_F} (a) and (b)). The convergence is still almost as smooth as that of RPROP. Hence, CoRe combines and improves the benefits of both, RPROP and Adam. We note that these trends also hold for a random training data selection (see Figures S7 (a) and (b) in the Supporting Information) instead of the lifelong adaptive data selection underlying the results in Figures \ref{fig:optimizer_E_F} (a) and (b).

%%%%%%%%%%%%%%%%%%%%%%%%%%%%%%%%%%%%%%%%%%%%%%%%%%%%%%%%%%%%%%%%%%%%%%%%%%%%%%%%%%%%%%%%%%%%%%%%%%%%%%%%
\subsection{Lifelong Adaptive Data Selection}
%%%%%%%%%%%%%%%%%%%%%%%%%%%%%%%%%%%%%%%%%%%%%%%%%%%%%%%%%%%%%%%%%%%%%%%%%%%%%%%%%%%%%%%%%%%%%%%%%%%%%%%%

\begin{figure}[htb!]
\centering
\includegraphics[width=\columnwidth]{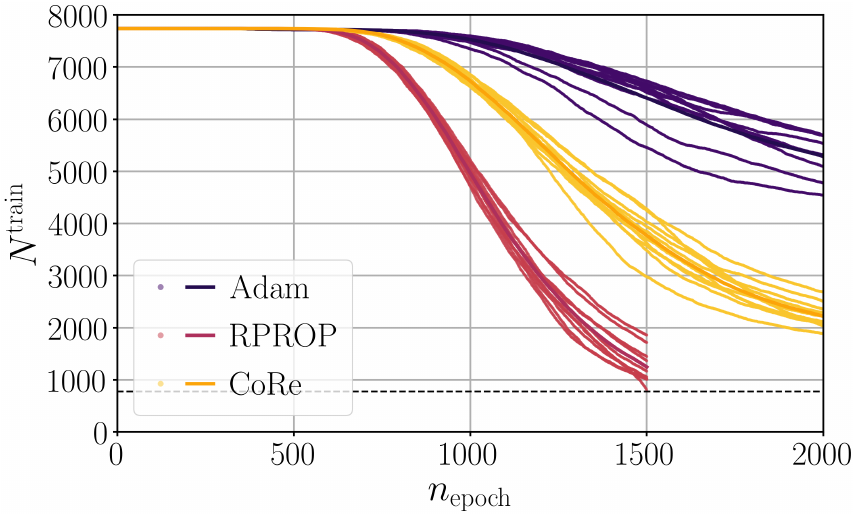}
\caption{Training data set reduction of the optimizers Adam, RPROP, and CoRe for training reference data set B. The number of considered training conformations $N^\mathrm{train}$ is shown as a function of the training epoch $n_\mathrm{epoch}$. The values of $N^\mathrm{train}$ of individual HDNNPs are represented by dots, while their mean is shown by a solid line. The black dashed line represents the number of training conformations which was used for fitting in each epoch.}\label{fig:optimizer_N}
\end{figure}

Continual data set reduction is essential for lifelong machine learning to keep the amount of training data for rehearsal on a manageable level during incremental learning of new data. Figure \ref{fig:optimizer_N} reveals how the training data set is narrowed during fitting. The training data reduction sets in after about 600 epochs because for data exclusion the adaptive selection factor $S_\mathrm{hist}^r$ needs to be decreased below $S_\mathrm{hist}^\mathrm{min}$ or increased above $S_\mathrm{hist}^\mathrm{max}$ by a specified number of consecutive applications of the decrease or increase factors, respectively (Algorithm 2). The optimization with RPROP plateaus already after about 600 epochs. Therefore, the data exclusion is too fast in the subsequent epochs, since the loss contribution of most conformations does not change much, biasing the importance evaluation by the adaptive selection factors. By contrast, the fluctuations in the convergence behavior of the optimizations with the Adam optimizer lead to a slow reduction of the training data. The reason for this is that the data importance measures also undergo the fluctuations which hamper to overcome the exclusion thresholds. The training data reduction of optimizations using CoRe is in between RPROP and Adam yielding a more balanced process. The number of excluded conformations per epoch also reduces for advanced training stages in later epochs making the training more stable (see also Figure S5 in the Supporting Information). By contrast, the optimizations with RPROP become unstable after about 1500 epochs due to the too rapid and strong data reduction.

In the training of reference data set B using the CoRe optimizer, the lifelong adaptive data selection assigned on average $(5.5\pm0.3)\cdot10^3$ training conformations to be redundant after 2000 epochs. Therefore, the training data was reduced to $29\%$ of the initial amount. Figures \ref{fig:optimizer_E_F} (a) and (b) show that this data reduction does not lead to a decline of the accuracy, which would be expected if training is performed on a non-representative subset of the training conformations. Despite the strong reduction of the amount of data, the lifelong adaptive data selection significantly improves the training accuracy compared to random data selection based on all training data (see Table S5 and Figures S7 (a) and (b) in the Supporting Information). This trend is observed for all optimizers. For the CoRe optimizer lifelong adaptive data selection yields an improvement of the ensemble RMSE values by $31\%$ for the energies and $48\%$ for the atomic force components compared to random data selection.

On average $38\pm11$ conformations were excluded from training because the model was not able to represent these conformations with a high accuracy. In this way, hindrance of the training process by these conformations can be avoided. Since the same 24 conformations were excluded in more than half of the training processes, these conformations are likely to be doubtful. Conformations, which are excluded only by a few individual HDNNPs of the ensemble, can still be predicted by the ensemble average (see Figures S6 (a) and (b) in the Supporting Information). In this way, the individual training processes can be improved, while the generalization of the ensemble prediction is still provided. Arising uncertainty for certain conformations due to their exclusion in some HDNNP trainings is covered by the uncertainty quantification. Therefore, this approach does not affect the reliability of the method.

%%%%%%%%%%%%%%%%%%%%%%%%%%%%%%%%%%%%%%%%%%%%%%%%%%%%%%%%%%%%%%%%%%%%%%%%%%%%%%%%%%%%%%%%%%%%%%%%%%%%%%%%
\subsection{Lifelong Machine Learning Potentials}
%%%%%%%%%%%%%%%%%%%%%%%%%%%%%%%%%%%%%%%%%%%%%%%%%%%%%%%%%%%%%%%%%%%%%%%%%%%%%%%%%%%%%%%%%%%%%%%%%%%%%%%%

Three frequently occurring example cases are explored, in which lifelong learning can be beneficial in comparison to iterative cycles of data set expansion and constructing new MLPs trained on all data. These cases represent training data completion of a sparsely sampled conformation space, expansion of the represented conformation space for the same chemical systems, and learning additional chemical systems. Lifelong learning can add an arbitrary number of new data points in each training epoch and does not have to be applied in a block-wise scheme as used in conventional active learning for MLPs. In this work, however, we added new data only in a single training epoch for a clear characterization of the resulting effects.

\begin{figure}[htb!]
\centering
\includegraphics[width=\columnwidth]{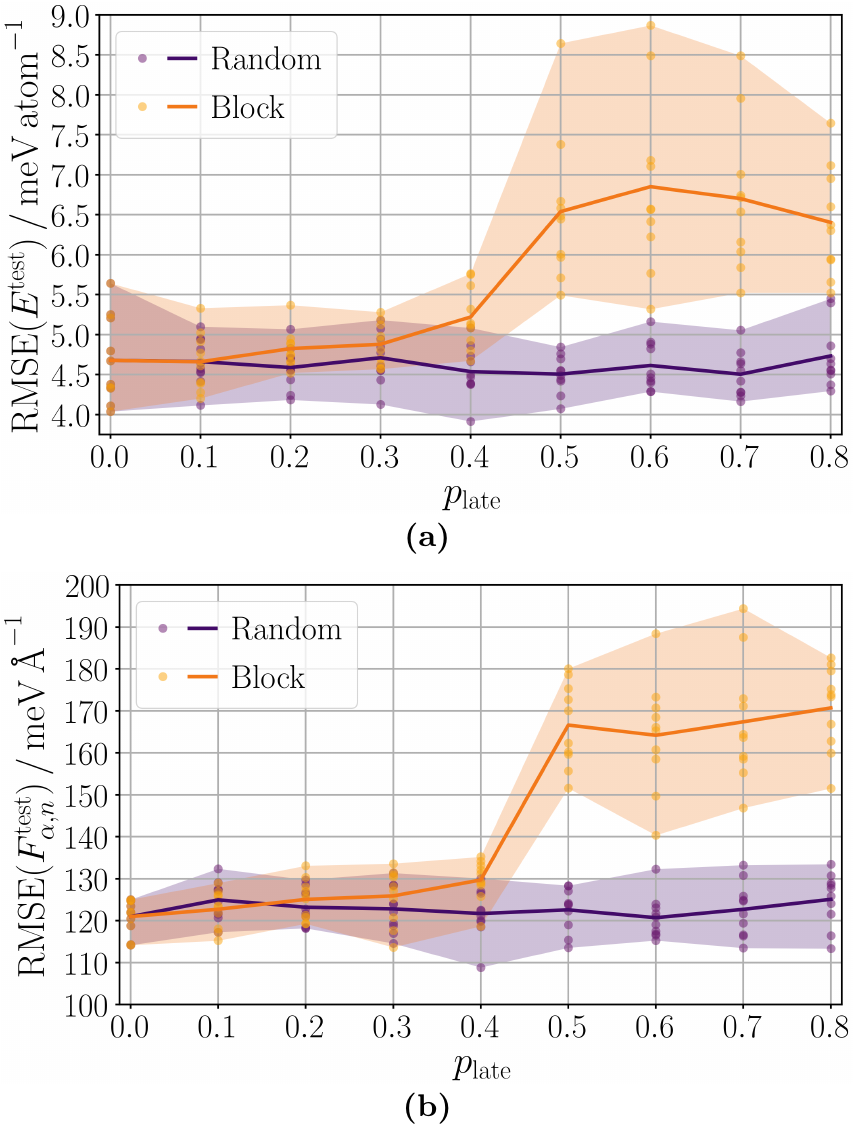}
\caption{Final accuracy of lMLPs for which a fraction of $p_\mathrm{late}$ training data of reference data set B was first available at a late training epoch. These data were either chosen randomly or a certain block was used. More detailed information about the procedure is provided in the main text. The test RMSE values of \textbf{(a)} energies $E^\mathrm{test}$ and \textbf{(b)} atomic force components $F_{\alpha,n}^\mathrm{test}$ are shown as a function of the late data fraction $p_\mathrm{late}$. RMSE values of individual HDNNPs are represented by dots, their mean by a solid line, and their range by a lighter colored band. The CoRe optimizer ($\beta_1^\mathrm{b}=0.7$ for ``Block'' with $p_\mathrm{late}>0$ and $\beta_1^\mathrm{b}=0.725$ otherwise) and lifelong adaptive data selection were applied for 1500 epochs.}\label{fig:lifelonglearning_E_F_final}
\end{figure}

A sparsely sampled conformation space was obtained for the initial training epochs when a high random fraction of the training data was first available at a late epoch. Figures \ref{fig:lifelonglearning_E_F_final} (a) and (b) show that the proposed lifelong learning strategies can handle this case very well yielding an almost constant final lMLP accuracy with respect to the late data fraction $p_\mathrm{late}$. The number of epochs, after which the late data fraction was added, was lower for high $p_\mathrm{late}$ because otherwise the small fraction of initial data would be overfitted. Figures S8 (b) and S9 in the Supporting Information show that the additional data were added at those epochs where the accuracy of the test atomic force components plateaus or even increases. The mean ensemble accuracy for $p_\mathrm{late}\in[0.5,0.8]$ is $\mathrm{RMSE}(\overline{E})=2.6\,\mathrm{meV\,atom}^{-1}$ and $\mathrm{RMSE}(\overline{F}_{\alpha,n})=68\,\mathrm{meV}\,\text{\AA}^{-1}$ and hence very similar to that for $p_\mathrm{late}=0$ (Table \ref{tab:optimizer_accuracy}). In addition to the flexibility gained in the training process, the lifelong learning approach requires less training data to be handled in the initial epochs compared to training on all data. Figure S9 in the Supporting Information reveals that the number of excluded conformations by the lifelong adaptive data selection is similar after 1500 epochs for different values of $p_\mathrm{late}$.

\begin{table}[htb!]
\caption{RMSE values of individual HDNNPs and the ensemble trained on reference data set C using learning on a stationary batch of all training data and lifelong learning. In lifelong learning, only the conformations of reference data set B were trained for the initial 1250 epochs and then the additional conformations of reference data set C were added. The CoRe optimizer with $\beta_1^\mathrm{b}=0.7$ and lifelong adaptive data selection were applied for 2500 epochs.}
\begin{center}
\begin{tabular}{lll}
\hline\vspace{-0.325cm}\\
Individual HDNNPs & Stationary & Lifelong\\
 & data & learning\vspace{0.075cm}\\
\hline\vspace{-0.325cm}\\
$\mathrm{RMSE}(E^\mathrm{train})\,/\,\mathrm{meV\,atom}^{-1}$ & $6.1\pm0.2$ & $6.9\pm0.6$\\
$\mathrm{RMSE}(E^\mathrm{test})\,/\,\mathrm{meV\,atom}^{-1}$ & $6.8\pm0.2$ & $7.8\pm0.5$\\
$\mathrm{RMSE}(F_{\alpha,n}^\mathrm{train})\,/\,\mathrm{meV}\,\text{\AA}^{-1}$ & $168\pm3$ & $185\pm12$\\
$\mathrm{RMSE}(F_{\alpha,n}^\mathrm{test})\,/\,\mathrm{meV}\,\text{\AA}^{-1}$ & $182\pm4$ & $205\pm10$\vspace{0.075cm}\\
\hline\vspace{-0.325cm}\\
Ensemble\vspace{0.075cm}\\
\hline\vspace{-0.325cm}\\
$\mathrm{RMSE}(\overline{E})\,/\,\mathrm{meV\,atom}^{-1}$ & $4.3$ & $4.5$\\
$\mathrm{RMSE}(\overline{F}_{\alpha,n})\,/\,\mathrm{meV}\,\text{\AA}^{-1}$ & $121$ & $122$\vspace{0.075cm}\\
\hline
\end{tabular}
\end{center}
\label{tab:lifelong_learning_accuracy}
\end{table}

To examine the performance for the case of the expansion of the represented conformation space for the same chemical systems, an lMLP was first trained on reference data set B for 1250 epochs. Subsequently, the additional structurally distorted conformations of reference data set C were added and training was continued for another 1250 epochs. Table \ref{tab:lifelong_learning_accuracy} reveals that lifelong learning yields RMSE values for individual HDNNPs which are about $13\%$ higher than those of learning on a stationary batch of all training data of reference data set C (see Figures S10 (a) and (b) and S11 in the Supporting Information for the training process). However, most of this lost accuracy is regained by the ensemble model which efficiently reduces the increased model variance (see Table \ref{tab:lifelong_learning_accuracy} and Figures S12 (a) and (b) in the Supporting Information for the prediction error distribution). Hence, the lMLP concept is able to extend the represented conformation space, while it retains the accuracy.

The higher RMSE values obtained in the training of reference data set C compared to B result from the even larger energy range and broader atomic force component distribution (see Table \ref{tab:reference_data} and Table S4 in the Supporting Information), while the model architecture and training hyperparameters remained unchanged. Still, especially the ensemble atomic force component RMSE is similar to other state-of-the-art HDNNPs trained for less elements \cite{Behler2021}.

To investigate the efficiency for learning additional chemical systems, the system-sorted reference data set B is split into two blocks, whereby the fraction of the second block is $p_\mathrm{late}$. The conformations are alphabetically sorted in the order central carbon atom, leaving group, and nucleophile. Learning additional S$_\text{N}$2 reactions for central \textit{tert}-butyl carbon atoms, i.e., adding nucleophiles and leaving groups at a late epoch which are only known for central methyl carbon atoms, yields a similar accuracy as learning on all data from the beginning ($p_\mathrm{late}<0.5$ in Figures \ref{fig:lifelonglearning_E_F_final} (a) and (b)). For $p_\mathrm{late}\geq0.5$ only reactions with central methyl carbon atoms are contained in the initial training data leading to an increase of the final test RMSE values. We emphasize that for $p_\mathrm{late}\geq0.8$ also some elements are missing in the initial training data. Still, the accuracy is better than that obtained using RPROP and for the energies it is similar to the Adam results (Table \ref{tab:optimizer_accuracy}).

Similar to the aforementioned case, ensembling can efficiently reduce the model variance introduced by incremental learning and hence is an important tool for lifelong machine learning. The ensemble accuracy is $\mathrm{RMSE}(\overline{E})=3.1\,\mathrm{meV\,atom}^{-1}$ and $\mathrm{RMSE}(\overline{F}_{\alpha,n})=78\,\mathrm{meV}\,\text{\AA}^{-1}$ for $p_\mathrm{late}\in[0.5,0.8]$ and therefore about $21\%$ larger than for $p_\mathrm{late}=0$. However, in these cases of learning additional chemical systems the lifelong learning strategies require further development to be on par with the accuracy of training on all data.

As this work is a proof of concept for lMLPs, more and different chemical systems need to be explored in future work to fine-tune and improve the lifelong learning strategies. Still, we showed that lifelong learning can reach the same accuracy as training on all data, while the flexibility of the training process was significantly increased.

%%%%%%%%%%%%%%%%%%%%%%%%%%%%%%%%%%%%%%%%%%%%%%%%%%%%%%%%%%%%%%%%%%%%%%%%%%%%%%%%%%%%%%%%%%%%%%%%%%%%%%%%
\subsection{Ensemble Prediction and Uncertainty Quantification}
%%%%%%%%%%%%%%%%%%%%%%%%%%%%%%%%%%%%%%%%%%%%%%%%%%%%%%%%%%%%%%%%%%%%%%%%%%%%%%%%%%%%%%%%%%%%%%%%%%%%%%%%

To validate the ensemble prediction and examine the uncertainty quantification, the lMLP trained by the CoRe optimizer on reference data set B was applied on conformations obtained from constrained DFT optimizations on a dense grid of $r_\mathrm{Cl-C}$ and $r_\mathrm{Br-C}$ distances for the S$_\mathrm{N}$2 reaction \ch{Br- + CH3Cl <=> BrCH3 + Cl-}. We emphasize that reference data set B uses a sparser and irregular sampling of this reaction and the validation conformations are unlikely to be in the reference data set B. Hence, the smoothness of the lMLP potential energy surface can be validated. The validation set contained 1062 conformations with maximal atomic force components of $16\,\mathrm{eV}\,\text{\AA}^{-1}$.

\begin{figure}[htb!]
\centering
\includegraphics[width=\columnwidth]{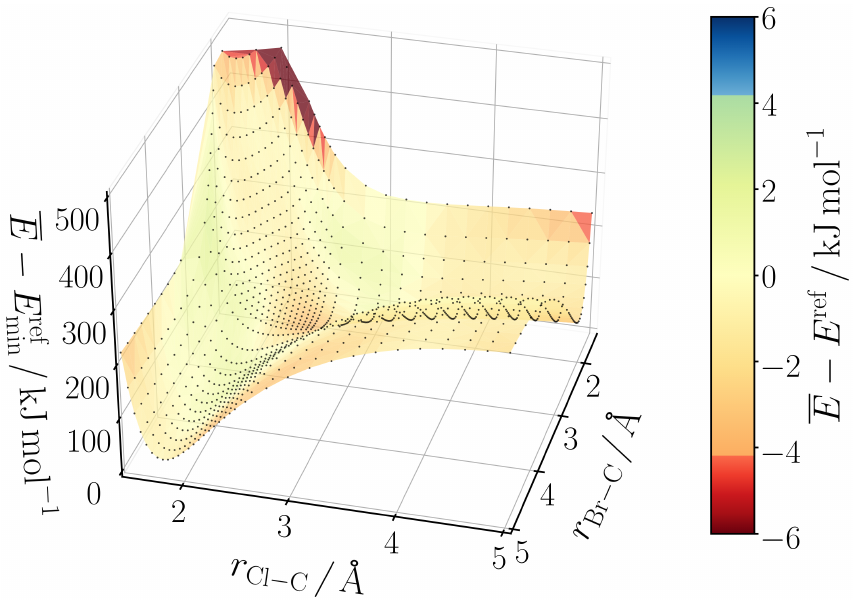}
\caption{Potential energy surface of the S$_\mathrm{N}$2 reaction \ch{Br- + CH3Cl <=> BrCH3 + Cl-}. The lMLP ensemble prediction energy $\overline{E}$ is referenced to the minimum DFT reference energy $E_\mathrm{min}^\mathrm{ref}$ of the given conformation space spanned by DFT optimized structures with constrained distances $r_\mathrm{Cl-C}$ and $r_\mathrm{Br-C}$. The color represents the error of $\overline{E}$ with respect to the DFT reference energy $E^\mathrm{ref}$. Black dots show the explicit evaluations of the lMLP. The colors between black dots are interpolated.}\label{fig:PES}
\end{figure}

Figure \ref{fig:PES} reveals that most sections of the represented potential energy surface are predicted within chemical accuracy, i.e., $1\,\mathrm{kcal\,mol}^{-1}=4.184\,\mathrm{kJ\,mol}^{-1}$, with respect to the DFT reference energies. For this six atom system an error of less than $7.2\,\mathrm{meV\,atom}^{-1}$ is therefore required. Larger errors are only observed for high energy conformations, which we expected because the maximal atomic force component can be $1\,\mathrm{eV}\,\text{\AA}^{-1}$ higher than that of the training data. The small errors with smooth distributions in the trained conformation space prove the smoothness of the lMLP.

\begin{figure}[htb!]
\centering
\includegraphics[width=\columnwidth]{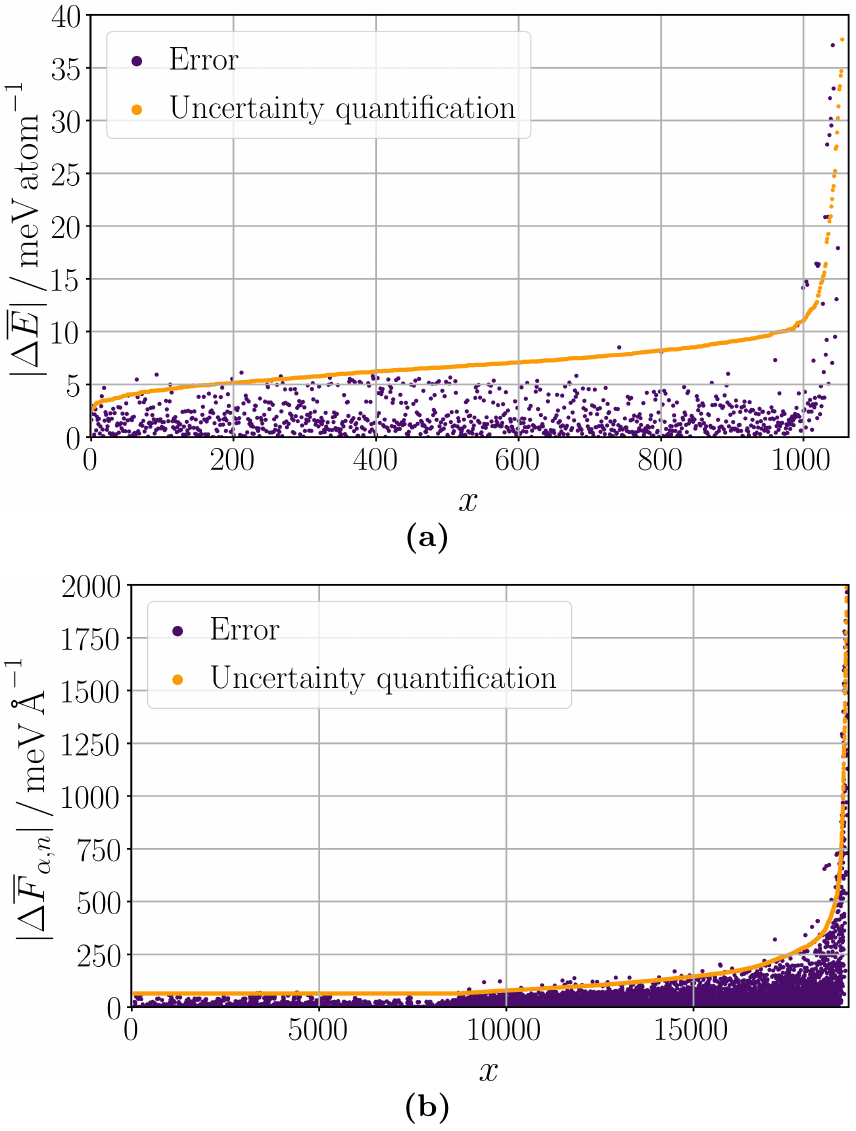}
\caption{Absolute values of the errors with respect to the DFT reference and the uncertainty quantification for the ensemble prediction of \textbf{(a)} energies $|\Delta\overline{E}|$ and \textbf{(b)} atomic force components $|\Delta\overline{F}_{\alpha,n}|$ of the validation data. The order $x$ sorts the validation conformations according to their uncertainty quantification.}\label{fig:uncertainty_E_F}
\end{figure}

The reliability of the uncertainty quantification is addressed in Figure \ref{fig:uncertainty_E_F} (a) and (b). The uncertainty quantification is equal to or larger than the absolute error with respect to the DFT reference energies for $98.6\%$ of the validation data with uncertainties $\Delta\overline{E}\leq10\,\mathrm{meV\,atom}^{-1}$. For the atomic force components the fraction is $99.7\%$ for $\Delta\overline{F}_{\alpha,n}\leq250\,\mathrm{meV}\,\text{\AA}^{-1}$. Hence, for most conformations the magnitude of the error is reliably predicted. For large errors the uncertainty quantification is expected to underestimate the errors (see Section 2.6) so that the above mentioned fractions decrease to $66\%$ for $\Delta\overline{E}>10\,\mathrm{meV\,atom}^{-1}$ and $93\%$ for $\Delta\overline{F}_{\alpha,n}>250\,\mathrm{meV}\,\text{\AA}^{-1}$. Still, a reliable identification of large errors is provided.

%%%%%%%%%%%%%%%%%%%%%%%%%%%%%%%%%%%%%%%%%%%%%%%%%%%%%%%%%%%%%%%%%%%%%%%%%%%%%%%%%%%%%%%%%%%%%%%%%%%%%%%%
\section{Conclusions}\label{sec:Conclusion}
%%%%%%%%%%%%%%%%%%%%%%%%%%%%%%%%%%%%%%%%%%%%%%%%%%%%%%%%%%%%%%%%%%%%%%%%%%%%%%%%%%%%%%%%%%%%%%%%%%%%%%%%

This work introduces the concept of lMLPs which can fine-tune and extend their representation in a rolling fashion. Hence, the lMLP concept unites MLP model efficiency and accuracy with flexibility. For an lMLP, a universal and computationally efficient atomic structure representation, an MLP model, uncertainty quantification, and lifelong learning strategies need to be combined.

Therefore, we introduced eeACSF vectors for the structural representation, which are size-independent with respect to the number of chemical elements in contrast to many other common MLP descriptors. Their representation performance of an S$_\text{N}$2 reference data set is similar to that of ACSF vectors at the break-even point of computational cost, which is at about four different elements. For several more elements eeACSF vectors become the only computationally reasonable option due to the combinatorial growth of ACSF vectors. An ensemble HDNNP model using eeACSFs can predict an S$_\text{N}$2 reference data set with ten different elements with a state-of-the-art accuracy of previous HDNNPs trained on less elements. Further, an ensemble of HDNNPs is a reliable way to quantify the uncertainty due to model variance and to identify conformations with high uncertainty in predictions. Additionally, ensembling increases the accuracy yielding potential energy surfaces with chemical accuracy for the S$_\text{N}$2 reactions.

As a basis of our lifelong learning strategies, we introduced the CoRe optimizer which can combine and improve the fast convergence of RPROP and the high final accuracy of the Adam optimizer. In the training of this work, the CoRe optimizer significantly improves the HDNNP ensemble accuracy by about $33\%$ for energies and forces compared to the Adam optimizer. Applying lifelong adaptive data selection further improves the accuracy and enables to narrow the training data set and exclude doubtful data during the training process. The CoRe optimizer and lifelong adaptive data selection can also improve training of machine learning models beyond lMLPs.

Finally, an lMLP can adapt to additional data which can be continuously added at any point in the training process. In this way, improvements of lMLPs are possible without learning again on all previous data and still a reliable method is obtained due to uncertainty quantification. In learning cases which are obtained during active learning or in the extension of the conformation space for the same reaction systems, the training accuracy is similar to that of learning on a stationary batch of all data. Even adding new reaction systems can be performed by the presented algorithms with only moderate accuracy loss. The benefit of lifelong learning is the enhanced flexibility of the training process enabling rolling explorations of chemical reactivity and training continuation of previous lMLPs. Moreover, adaptability of lMLPs is especially advantageous for large reference data sets where training on all data at once is computationally very demanding. We emphasize that the lMLP concept can also be applied for other MLP approaches beyond HDNNPs in future work.

\section*{Acknowledgement}

This work was supported by an ETH Zurich Postdoctoral Fellowship.

%\section*{Supporting Information}
%
%Activation function, weight initialization, DFT free atom energies, ACSF parameters, performance evaluation of activation function and weight initialization, energy distribution of reference data, prediction error distributions, training process comparison of ACSFs and eeACSFs, performance evaluation of random data selection, and lMLP training processes (PDF file).

%\bibliography{Bibliography}

\end{document}